%% file: main.tex
\documentclass[final,3p,times]{elsarticle}
\usepackage[numbers]{natbib}
\usepackage{hyperref}
\usepackage{comment}
\usepackage{bbm}
\usepackage{eucal}
\usepackage{amsmath, xparse}
\usepackage{here}
\usepackage{xcolor}
\usepackage{subfig}
\usepackage{arydshln}
\usepackage{algorithm}
\usepackage{algorithmic}
\mathchardef\mhyphen="2D

\usepackage{amssymb}

\journal{Pattern Recognition}

\begin{document}

\begin{frontmatter}

\title{
Graph Classification via Discriminative Edge Feature Learning 
}

\author[inst1]{Yang Yi}
\affiliation[inst1]{organization={School of Information Technology, Deakin University},
            city={Waurn Ponds},
            postcode={Victoria 3216}, 
            country={Australia}}

\ead{yang.yi@deakin.edu.au}

\author[inst1]{Xuequan Lu\corref{mycorrespondingauthor}}
\cortext[mycorrespondingauthor]{Corresponding author}
\ead{xuequan.lu@deakin.edu.au}

\author[inst1]{Shang Gao}
\ead{shang.gao@deakin.edu.au}

\author[inst1]{Antonio Robles-Kelly}
\ead{antonio.robles-kelly@deakin.edu.au}

\author[inst2]{Yuejie Zhang}
\affiliation[inst2]{organization={School of Computer Science, Fudan University},
            city={Shanghai},
            postcode={200433}, 
            country={China}}
\ead{yjzhang@fudan.edu.cn}

\begin{abstract}

Spectral graph convolutional neural networks (GCNNs) have been producing encouraging results in graph classification tasks. However, most spectral GCNNs utilize fixed graphs when aggregating node features, while omitting edge feature learning and failing to get an optimal graph structure. Moreover, many existing graph datasets do not provide initialized edge features, further restraining the ability of learning edge features via spectral GCNNs. In this paper, we try to address this issue by designing an edge feature scheme and an add-on layer between every two stacked graph convolution layers in GCNN. Both are lightweight while effective in filling the gap between edge feature learning and performance enhancement of graph classification. The edge feature scheme makes edge features adapt to node representations at different graph convolution layers. The add-on layers help adjust the edge features to an optimal graph structure. To test the effectiveness of our method, we take Euclidean positions as initial node features and extract graphs with semantic information from point cloud objects. The node features of our extracted graphs are more scalable for edge feature learning than most existing graph datasets (in one-hot encoded label format). Three new graph datasets are constructed based on ModelNet40, ModelNet10 and ShapeNet Part datasets. Experimental results show that our method outperforms state-of-the-art graph classification methods on the new datasets by reaching $96.56\%$ overall accuracy on Graph-ModelNet40, $98.79\%$ on Graph-ModelNet10 and $97.91\%$ on Graph-ShapeNet Part. The constructed graph datasets will be released to the community.

\end{abstract}


\end{frontmatter}

\input{paper/introduction}
\input{paper/relatedwork}

\input{paper/method}

\input{paper/results}

\input{paper/conclusion}

\bibliographystyle{elsarticle-num}
\bibliography{egbib}

\end{document}

%% file: paper/introduction.tex
\section{Introduction}

\label{sec:introduction}

In recent years, graph neural networks (GNNs) have been making advances in tasks such as graph classification, node classification and link prediction. For example, benefiting from the diversity and generality of graph representation, GNNs have been applied to biochemistry \cite{duvenaud2015convolutional,kearnes2016molecular,fout2017protein,de2018molgan,you2018graph}, recommendation system \cite{ying2018graph,wang2019knowledge,zhang2020personalized}, computer vision \cite{simonovsky2017dynamic,te2018rgcnn,wang2019dynamic,xu2020grid}, natural language processing \cite{peng2017cross,beck2018graph,marcheggiani2018exploiting} and many other fields \cite{luceri2019infringement,shin2019lifelog,valsesia2019image}. Since graph is a special manifold structure that does not possess translation invariance property, traditional convolutional neural networks (CNNs) cannot be directly used on graphs. Earlier works \cite{micheli2009neural,atwood2016diffusion,niepert2016learning} are based on spatial domain's GCNNs and support the transformation from non-Euclidean data to Euclidean data by fixing the number and order of nodes of graphs. Further, they accomplish the message propagation by constructing node feature aggregation function. 

In contrast, spectral GCNNs \cite{bruna2013spectral,estrach2014spectral,defferrard2016convolutional,zhang2018end,xu2019graph} perform feature aggregation for neighbouring nodes by constructing the convolution operators on graphs. The third-generation graph convolution layer \cite{kipf2016semi}, which is the most commonly used today, is simplified from the first two generations \cite{estrach2014spectral,defferrard2016convolutional} and has yielded encouraging results on node classification and graph classification. However, the graph convolution layer disregards edge feature learning while aggregating node features, thus limiting the learning ability of spectral GCNNs at the graph structure level.

Moreover, most commonly used graph classification datasets \cite{debnath1991structure,dobson2003distinguishing,borgwardt2005protein,wale2008comparison} provide graphs without edge features, which prevents the spectral GCNNs from exploiting graph structure information when applied in graph-level tasks. Node features of graphs are normally represented by one-hot encoded labels in most existing graph classification datasets. This significantly constrains spectral GCNNs from using these features in a more flexible way. Some graph-based point cloud classification networks \cite{simonovsky2017dynamic,te2018rgcnn,wang2019dynamic,xu2020grid} take the positions of point clouds in Euclidean space as node features and use the Euclidean distance between nodes to compute edge features. This method greatly promotes the scalability of both the node and edge features. The results in \cite{defferrard2016convolutional} also show that implementing Euclidean data on GCNNs are as feasible as classical CNN approaches.

In this paper, we design a novel edge feature scheme and an add-on layer between every two stacked graph convolution layers to better exploit the relationship between the connected nodes of graph. Unlike the default uniform weights (e.g., 1.0 for each edge), edge features are calculated by edge feature scheme in a weighted manner. The add-on layer takes the hidden representation of each node from the previous graph convolution layer and generates a matrix for updating the calculated edge features. In this way, discriminative edge features can be produced and updated, further facilitating the graph feature learning at the next graph convolution layer. We use graphs derived from point cloud objects to examine our approach due to the scalability of point clouds' node and edge features in Euclidean space.

Specifically, graphs are constructed by extracting a key node from each part (or cluster) of the point cloud, where the semantic information is encoded. To obtain part labels, we register the source point cloud (without part labels) with a target template point cloud (with part labels) and assign the point in the source with the label of the template counterpart that is the nearest point to that source point. Our approach is validated on ModelNet40 \cite{wu20153d}, ModelNet10 \cite{wu20153d} and ShapeNet Part \cite{yi2016scalable} datasets. Results show that our approach effectively improves the learning capability of the graph convolution layer and outperforms other state-of-the-art graph classification methods on our constructed graph datasets, reaching $96.56\%$ overall accuracy on ModelNet40 \cite{wu20153d}, $98.79\%$ on ModelNet10 \cite{wu20153d}, and $97.91\%$ on ShapeNet Part dataset \cite{yi2016scalable}.

Our main contributions are summarized as follows. 
\begin{itemize}
    
    \item We propose novel graph classification method for better classification performance.
    
    \item We develop an edge feature scheme and an add-on layer for discriminative feature learning. 
    
    \item We create corresponding graph datasets based upon ModelNet40, ModelNet10 and ShapeNet Part datasets and will release them to the community. 
    
    \item State-of-the-art classification performances are achieved on our generated graph datasets.
    
\end{itemize}

%% file: paper/relatedwork.tex
\section{Related Work}
\label{sec:relatedwork}

\subsection{Graph Convolutional Neural Networks}
\label{sec2:gnn}
Graph Convolutional Neural Networks (GCNNs) have been applied in multiple tasks, e.g., graph regression, classification and generation, node classification, regression and edge prediction. 
There are abundant graph datasets available for biochemistry \cite{borgwardt2005protein,debnath1991structure,wale2008comparison}, citation network \cite{giles1998citeseer,mccallum2000automating,sen2008collective} and social networking \cite{agarwal2009social,hamilton2017inductive,richardson2003trust}. Although graph-based classification methods mentioned in Sec. \ref{sec:introduction} constructed graphs for point clouds, there still lack unified graph datasets for \cite{wu20153d}.
According to the way of defining the graph convolution operator, GCNNs can be divided into two categories as follows:

\textbf{Spatial methods} focus on the node domain by aggregating the features of each central node and its neighbouring nodes. E.g., GAT \cite{velivckovic2017graph} mainly focused on learning from nodes instead of structure. It defined feature aggregation functions depending on the attention mechanism, and the edge feature was replaced by a normalized attention coefficient between nodes. GraphSAGE \cite{hamilton2017inductive} borrowed the idea of mini-batch to perform random sampling of neighbouring nodes. An aggregation function was also used to update the global features of nodes, thus giving the model the ability of learning inductively.

\textbf{Spectral methods} complete feature aggregation between nodes in the spectral domain. Spectral CNN \cite{bruna2013spectral} was the first proposal that constructed a spectral convolutional neural network with convolutional operators on graphs. Based on the convolution theorem, it realized the graph convolution and information aggregation between nodes. GCN \cite{kipf2016semi} simplified the Chebyshev network \cite{estrach2014spectral}, where a first-order GCNN was implemented and realized for applications in semi-supervised learning scenarios. Focusing on the graph classification task, Zhang et al. \cite{zhang2018end} implemented GCN and added a sort pooling layer to form a DGCNN model. It took the sort pooling layers to solve the standard problem of sorting vertices. The formula derivation showed that the output order of the sort-pooling layer was consistent with the structural roles of the nodes.

\subsection{Point Cloud Part Segmentation}
\label{sec2:part segmentation}
Supervised part segmentation techniques \cite{lei2020spherical,liu2020fg,lyu2020learning,thomas2019kpconv,xu2020learning} for point clouds have been evolving rapidly due to the increasing availability of human-annotated datasets \cite{mo2019partnet,yi2016scalable}. Unlike the supervised methods taking fixed annotations for segmented parts, semi-supervised and unsupervised methods provide ways for part representation learning. CoSegNet \cite{zhu2019cosegnet} took the unsegmented shapes in ComplementMe dataset \cite{sung2017complementme} as training input to create part collections. The output was K-way labelled objects, where K denoted the upper bound of the part numbers. BAE-NET \cite{chen2019bae} implemented a branched autoencoder network that minimized the loss by reconstructing the inside-outside status between points and shapes to obtain the simplest representations of part collections. PartNet \cite{yu2019partnet} output an unfixed part number for each class based on the complexity of the object’s geometric structure. RIMNet \cite{niu2022rim} recursively decomposed the point cloud into two parts and output hierarchical shape structures. It was the first fully unsupervised segmentation method without requiring any ground-truth segmented objects. Although all of these Co-Segmentation methods segment point clouds automatically, they all require separate training for a particular class and cannot achieve user-defined segmentation results.

%% file: paper/method.tex
\section{Method}
\label{sec3:method}

In this section, we firstly explain the method to constructing the graph from a point cloud object. Then, we show how our spectral GCNN is designed to better facilitate the classification task.

\subsection{Semantic Graph Construction}
\label{sec3:p1:graph construction}
Parts of a point cloud (e.g., two wings of an airplane) provide semantic information, which usually represent a common characteristic in the same point cloud class and can effectively help with the classification task. Based on this, points that are selected from those parts are discriminative, and further simple operations on these points are available to generate a graph. Therefore, we divide the semantic graph construction task into two steps: part annotation generation and graph extraction.

\subsubsection{Part annotation generation}
\label{sec3:p1.1:ICP}
ICP algorithm (iterative closest point) \cite{besl1992method} is chosen to generate the part annotation. We select at least one template point cloud with part labels for each point cloud category, and these template point clouds can be easily obtained by selecting a few point clouds from the target dataset and annotate their parts manually. Provided this, we can encode semantic part information by registering the template point cloud with those unlabelled point clouds.
Given an unlabeled point cloud as source $P_{s}$ and a template point cloud as target $P_{t}$, the nearest neighbours of $P_{s}$ in $P_{t}$ as $P_{t}^{s}$, and point $p_{s} \in P_{s}$, $p_{t} \in P_{t}$ and $p_{t}^{s} \in P_{t}^{s}$, the point cloud registration problem can be solved by minimizing the following function $\mathbb{E}$:
\begin{equation}\label{eq:ICP}
\mathbb{E}(P_{s},P_{t})= {\min} \frac{1}{\left|P_{s}\right|} \sum_{s=1}^{\left|P_{s}\right|}\left\|p_{t}^{s}-\left(R  p_{s}+T\right)\right\|^{2},
\end{equation}
where $R$ is a rotation matrix and $T$ is a translation vector. Both of them can be calculated with Singular Value Decomposition (SVD). After certain iterations or reaching the convergence on $\mathbb{E}$, the part label of each point in $P_{t}^{s}$ will be allocated to points in $P_{s}$ accordingly. This allows us to encode the semantic part information for the unlabeled point cloud.

\subsubsection{Graph extraction}
\label{sec3:p1.2:graph extraction}
After the part information generation step, each point in the source point cloud has a part label now. Nevertheless, a part may have multiple sub-parts that exist independently in 3D space. For instance, a table has four separate table legs. To further distinguish these sub-parts, we compute distances among all points within each part $P$, and record the adjacency relationship (e.g., 0 for non-adjacent and 1 for adjacent) of these points in a neighborhood matrix $\mathbf{N}_{i,j}$ by comparing all distances with a threshold $\tau$. Where $\{\mathbf{N}_{i,j}| i\in [1,\mathcal N_{p}], j\in [1,\mathcal N_{p}]\}$, $\mathcal N_{p}$ denotes the point number of the annotated part $P$. 

Furthermore, we implement the simple yet effective Depth-First-Search (DFS) on $\mathbf{N}_{i,j}$, each time DFS returns all the connected points that constitute the sub-part, and these connected points are stored before being removed from $P$. This searching process repeats until the number of remaining points in $P$ becomes 0. After being separated, the sub-parts will replace the original part and act as new parts in the point cloud. We extract a node from each part by averaging all points of that part, and its initial node feature is the averaged position of that part. This representation greatly simplifies the original point cloud, as well as effectively encodes the semantic part information.

Suppose $\mathbf{G}=\{\mathbf{V}, \mathbf{E}\}$ represents a graph, where $\mathbf{V}$ is the collection of all nodes, $\mathbf{E}$ is the collection of all edges, $\mathrm{F}^{V}$ and $\mathrm{F}^{E}$ denote the node features and edge features of the graph $\mathbf{G}$, respectively. The adjacency matrix $\mathbf{A}$ can be easily computed over $\mathbf{G}$. To facilitate the follow-up training, we bidirectionally connect all nodes extracted from the source point cloud to construct a complete directed graph.

\begin{figure}
\centering
\includegraphics[width=1.0\textwidth]{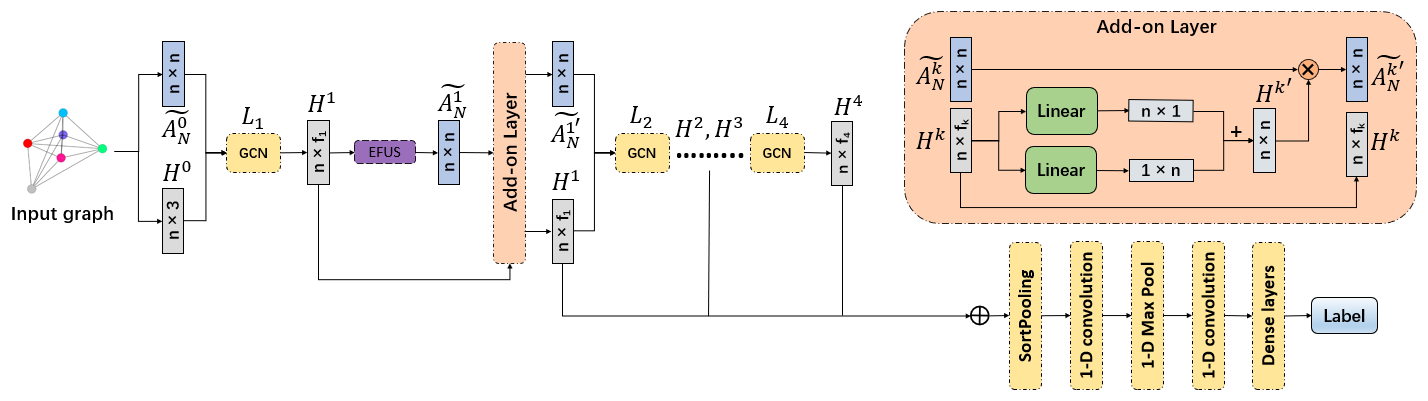}
\caption{Architecture of our graph classification network. Three add-on layers and edge feature schemes are added between the four stacked graph convolution layers. In each add-on layer, the dimension of $\mathbf{H}^{k}$ is deducted by two linear layers separately. The output row and column vectors are broadcast added to produce the update matrix for each normalized edge feature. $\mathbf{H}^{k}$ and the updated adjacency matrix $\tilde{\mathbf{A}}_{N}^{{k}^{\prime}}$ are then fed into the next graph convolution layer. $\oplus$ means the concatenation of all $\mathbf{H}^{k}$.}
\label{fig:ch3:architecture}
\end{figure}

\subsection{Graph Classification}
\label{sec3:p2:graph classify}

\subsubsection{Network architecture}
\label{sec3:p2.1:Architecture}
The architecture of our graph classification network is shown in Fig. \ref{fig:ch3:architecture}. DGCNN \cite{zhang2018end} is adopted as the backbone model in our method. We choose row normalized adjacency matrix as our feature aggregation function for node correlation, given its outstanding performance in graph learning tasks \cite{atamna2019spi,kampffmeyer2019rethinking,zhang2018end}. Four stacked graph convolution layers are implemented to learn node and edge features of graphs.  The hidden representation matrix $\mathbf{H}^{k}\in \mathbb{R}^{n \times f_{k}}$ at layer $k$ as:
\begin{equation}\label{eq:hk}
\mathbf{H}^{k}=\delta\left(\tilde{\mathbf{D}}^{-1} \tilde{\mathbf{A}} \mathbf{H}^{k-1} \mathbf{W}^{k-1}\right),
\end{equation}
where $n$ is the node number of a graph and $f_{k}$ is the dimension of each node feature. For each layer $k$, We employ the same $f_{k}$ as in \cite{zhang2018end} which are 32, 32, 32 and 1, respectively. $\mathbf{H}^{0}\in \mathbb{R}^{n \times 3}$, given each node takes its Euclidean position as the origin graph signal. $\tilde{\mathbf{A}}\in \mathbb{R}^{n \times n}$ denotes the self-loop added adjacency matrix, $\tilde{\mathbf{D}}\in \mathbb{R}^{n \times n}$ is the diagonal degree matrix of $\tilde{\mathbf{A}}$, and $\delta$ is the non-linear activation function. $\mathbf{W}^{k-1} \in \mathbb{R}^{f_{k-1} \times f_{k}}$ is a trainable convolution parameter. Between each of the two consecutive layers, an edge feature scheme and an add-on layer are added to enhance the network's node and edge features learning ability. Specified discussions are made in the remaining part of this section.

These four hidden representations are then concatenated, sorted and pooled in the SortPooling layer, where the first $z$ rows are retained. Afterwards, traditional MaxPooling layers, 1-D convolutional layers and Dense layers are used for prediction.

\subsubsection{Edge feature scheme}
\label{sec3:p2.2:Edge features}
Edge features are important in graph learning. By default, all input edge features in $\mathrm{F}^{E}$ are set to 1, which can hardly reflect the graph structure effectively. Besides of that, edge features are normally fixed across all graph convolution layers. The default setting does not match the corresponding node features that are updated during the node feature aggregation at each graph convolution layer. To address this problem, we define a scheme to dynamically learn edge features in the network.

For input edge features, an intuitive choice is a Gaussian-like function that can generate a ``weight'' as the edge feature based on the distance between two connected nodes. 
\begin{equation}
\label{eq:gmm}
\mathrm{F}^E(x) =\frac{1}{(2 \pi)^{1 / 2} \sigma}
e^{\left[-\left(\frac{x-\mu}{2\sigma}\right)^{2}\right]}, 
\end{equation}
where $x$ is the Euclidean distance between two nodes. $\mu$ and $\sigma$ are constants, representing the mean value and standard deviation of node distances among all graphs, respectively. We also extend this function to compute the edge features between every two consecutive graph convolution layers. $x$ becomes the $L_{2}\mhyphen norm$ of any two node representations in high dimensions. The new edge features are expected to be more adaptive to the current node representations, further facilitating graph learning before entering the next graph convolution layer.

Since this function has a negative sign in the exponent, the result will be lower if the distance between nodes becomes larger, and vice versa. According to the semantics of a point cloud, closer nodes are more likely to have similar part information than far-apart nodes. Thus, assigning a smaller value to the edge with a larger distance may result in inaccurate topological information. On the contrary, a stronger connection (i.e., larger edge feature) between nodes from far-apart parts will benefit the encoding of global topological information, which may help produce a better graph structure. Further discussion will be made in Sec. \ref{sec4:p5:edge features} on the choice of design.

Based upon these observations, we define the edge features in adjacency matrix $\tilde{\mathbf{A}}$ at layer $k$ with the following exponential function, where ${v}_{w}\in \mathbf{V}$, ${v}_{q}\in \mathbf{V}$, $\mathrm{H}^{k}_{{v}_{w}}\in \mathbb{R}^{f_{k}}$ and $\mathrm{H}^{k}_{{v}_{q}}\in \mathbb{R}^{f_{k}}$:

\begin{equation}\label{eq:+x}
\tilde{\mathbf{A}}^{k}_{w,q} = e^{\left\|\mathrm{H}^{k}_{{v}_{w}} - \mathrm{H}^{k}_{{v}_{q}}\right\|_{2}},
\end{equation}

\subsubsection{Add-on layer}
\label{sec3:p2.3:add-on layer}

Although the edge feature scheme enables the calculation of edge features with the corresponding node representations, optimized edge features are still pursued to better help with the node feature aggregations in the following graph convolution layer. To address this problem, we design a mechanism to update the $\tilde{\mathbf{A}}^{k}$ with learned features through the network.

Suppose a normalized adjacency matrix at layer $k$ is defined  as $\tilde{\mathbf{A}}_{N}^{k}$, where $\tilde{\mathbf{A}}_{N}^{k} = \tilde{\mathbf{D }}^{-1} \tilde{\mathbf{A}}^{k}$, then a novel add-on layer is added after the edge feature scheme to update the edge features in $\tilde{\mathbf{A}}_{N}^{k}$. This specific procedure is depicted in the top-right add-on layer block in Fig. \ref{fig:ch3:architecture}. It can be represented by:

\begin{equation}\label{eq:H adjust}
\mathbf{H}^{{k}^{\prime}}= \mathrm{L}_1\left(\mathbf{H}^{k}\right) + \left(\mathrm{L}_2\left(\mathbf{H}^{k}\right)\right)^\intercal,
\end{equation}

\begin{equation}\label{eq:A adjust}
\tilde{\mathbf{A}}_{N}^{k^{\prime}} = \mathbf{H}^{{k}^{\prime}} \times \tilde{\mathbf{A}}_{N}^{k},
\end{equation}

\begin{equation}\label{eq: H update}
\mathbf{H}^{k+1} = \delta\left(\tilde{\mathbf{A}}_{N}^{k^{\prime}} \mathbf{H}^{k} \mathbf{W}^{k}\right),
\end{equation}

In Eq. \eqref{eq:H adjust}, $\mathrm{L}_1$ and $\mathrm{L}_2$ represent two linear layers, and both their output channels are set to 1. The symbol $+$ denotes the broadcast addition of matrices. Eq. \eqref{eq:A adjust} is used for updating the $\tilde{\mathbf{A}}_{N}^{k}$ through matrix multiplication. In Eq. \eqref{eq: H update}, the updated $\tilde{\mathbf{A}}_{N}^{k^{\prime}}$ is subsequently used for computing the hidden representation $\mathbf{H}^{k+1}$.

%% file: paper/results.tex
\section{Experiments} 
\label{sec4:results}

\subsection{Constructed Graph Datasets}
\label{sec4:p1:dataset}

\begin{figure}[t]
\centering
\begin{minipage}[b]{0.49\linewidth}
    \begin{minipage}[b]{0.47\linewidth}
        \begin{minipage}[b]{0.47\linewidth}
        {\label{}\includegraphics[width=1\linewidth]{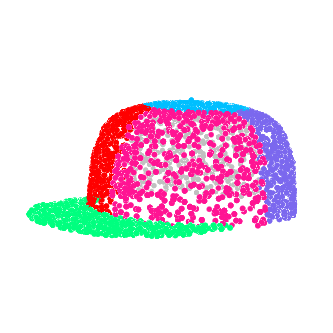}}
        \end{minipage}
        \begin{minipage}[b]{0.47\linewidth}
        {\label{}\includegraphics[width=1\linewidth]{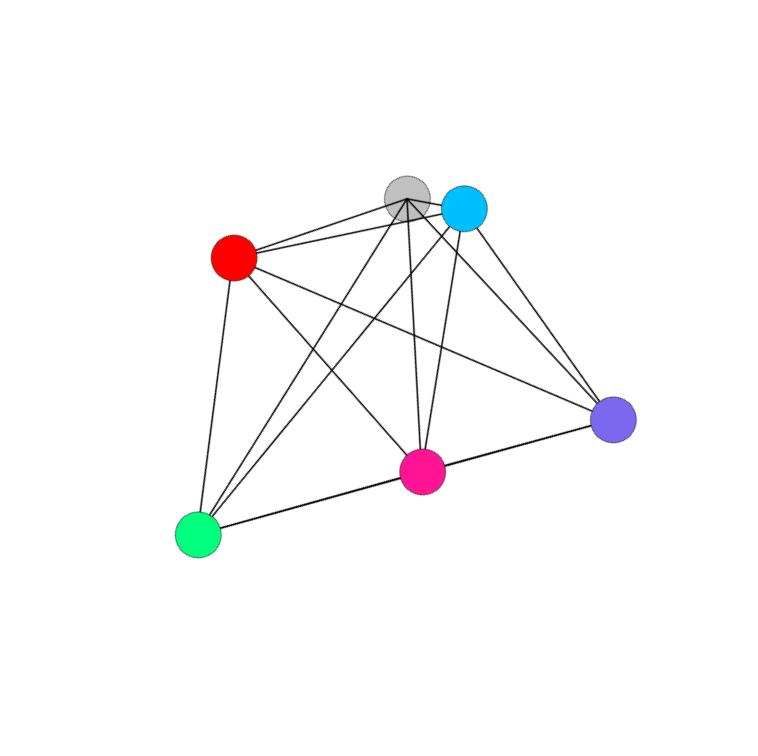}}
        \end{minipage}
    \centerline{\scriptsize{Cap}}
    \end{minipage}
    \begin{minipage}[b]{0.47\linewidth}
        \begin{minipage}[b]{0.47\linewidth}
        {\label{}\includegraphics[width=1\linewidth]{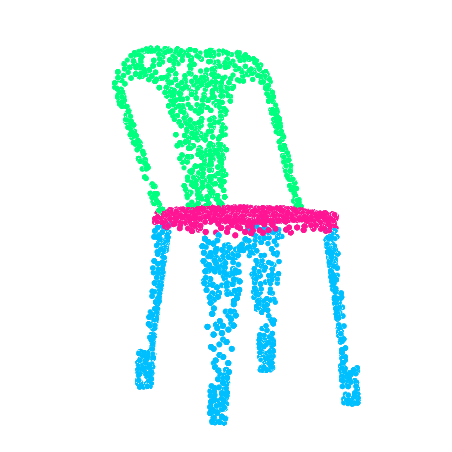}}
        \end{minipage}
        \begin{minipage}[b]{0.47\linewidth}
        {\label{}\includegraphics[width=1\linewidth]{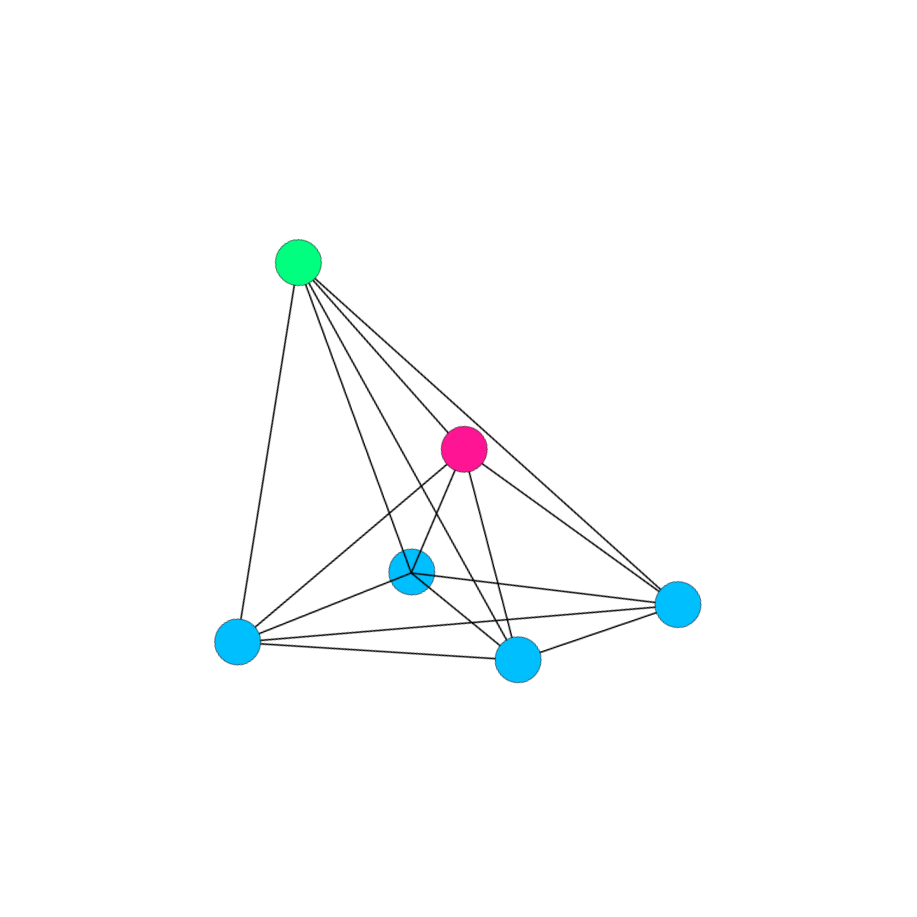}}
        \end{minipage}
    \centerline{\scriptsize{Chair}}
    \end{minipage}
\end{minipage}
\begin{minipage}[b]{0.49\linewidth}
    \begin{minipage}[b]{0.47\linewidth}
        \begin{minipage}[b]{0.47\linewidth}
        {\label{}\includegraphics[width=1\linewidth]{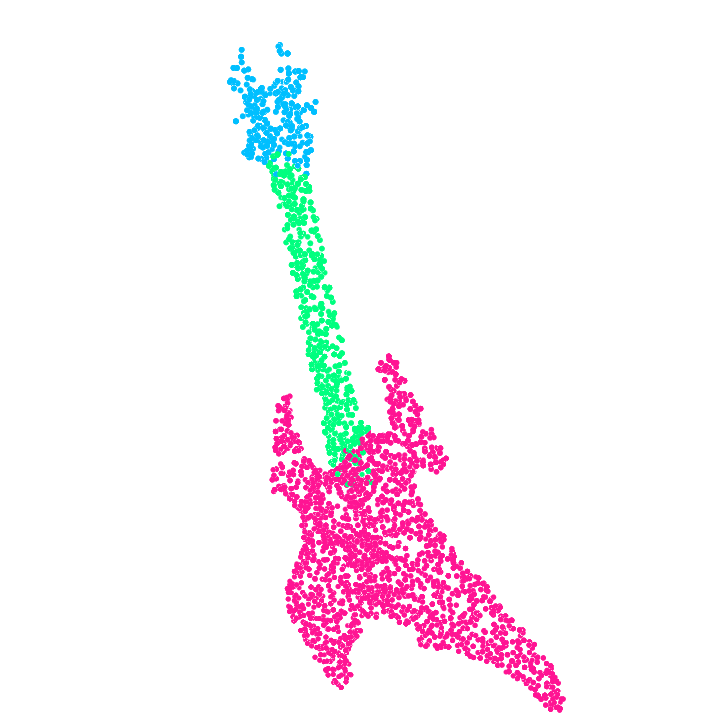}}
        \end{minipage}
        \begin{minipage}[b]{0.47\linewidth}
        {\label{}\includegraphics[width=1\linewidth]{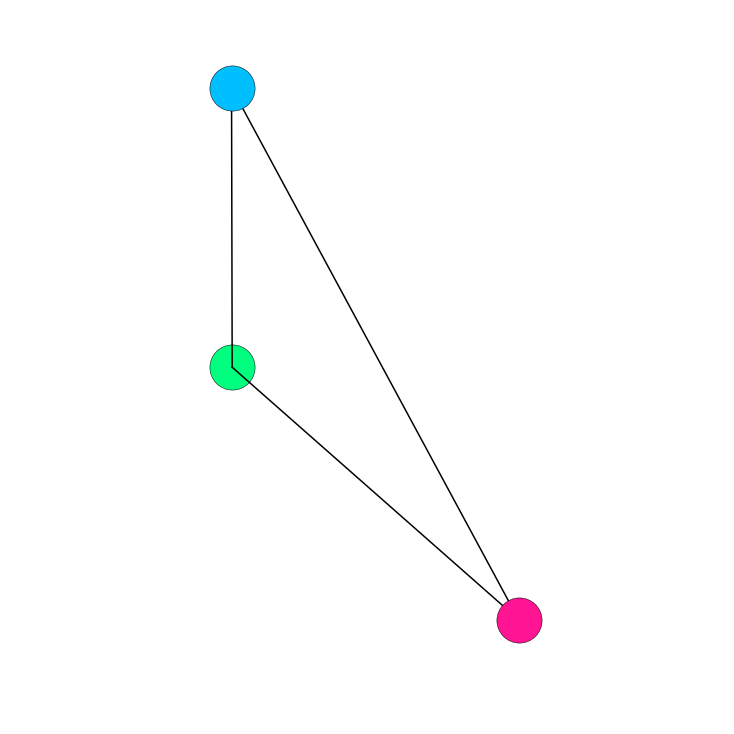}}
        \end{minipage}
    \centerline{\scriptsize{Guitar}}
    \end{minipage}
    \begin{minipage}[b]{0.47\linewidth}
        \begin{minipage}[b]{0.47\linewidth}
        {\label{}\includegraphics[width=1\linewidth]{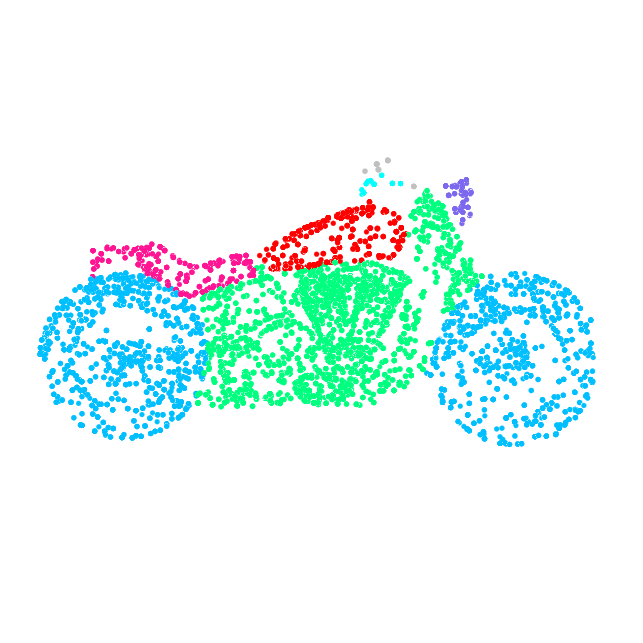}}
        \end{minipage}
        \begin{minipage}[b]{0.47\linewidth}
        {\label{}\includegraphics[width=1\linewidth]{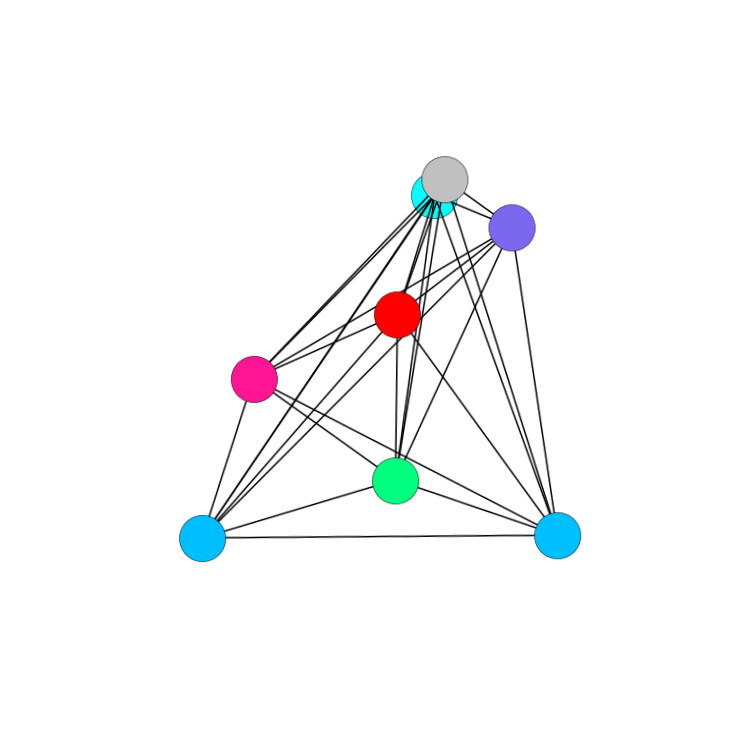}}
        \end{minipage}
    \centerline{\scriptsize{Motorbike}}
    \end{minipage}
\end{minipage}
\\
\begin{minipage}[b]{0.49\linewidth}
    \begin{minipage}[b]{0.47\linewidth}
        \begin{minipage}[b]{0.47\linewidth}
        {\label{}\includegraphics[width=1\linewidth]{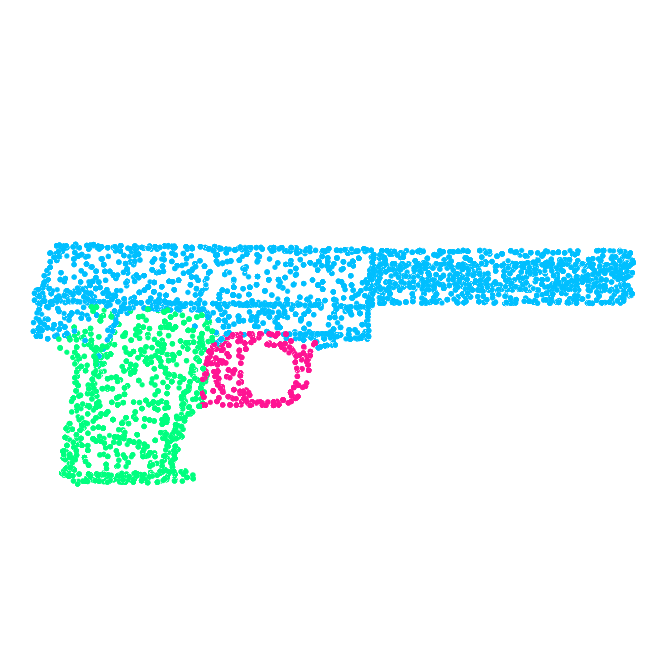}}
        \end{minipage}
        \begin{minipage}[b]{0.47\linewidth}
        {\label{}\includegraphics[width=1\linewidth]{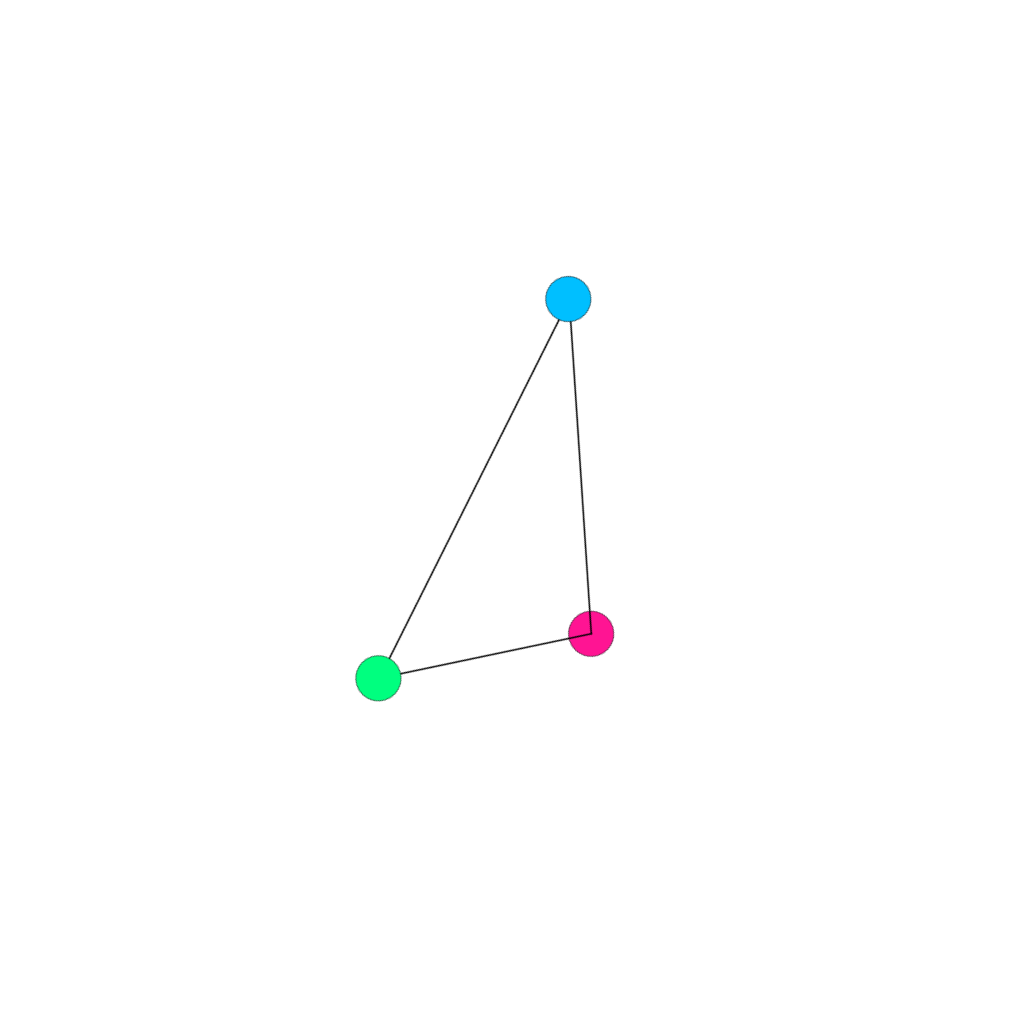}}
        \end{minipage}
    \centerline{\scriptsize{Pistol}}
    \end{minipage}
    \begin{minipage}[b]{0.47\linewidth}
        \begin{minipage}[b]{0.47\linewidth}
        {\label{}\includegraphics[width=1\linewidth]{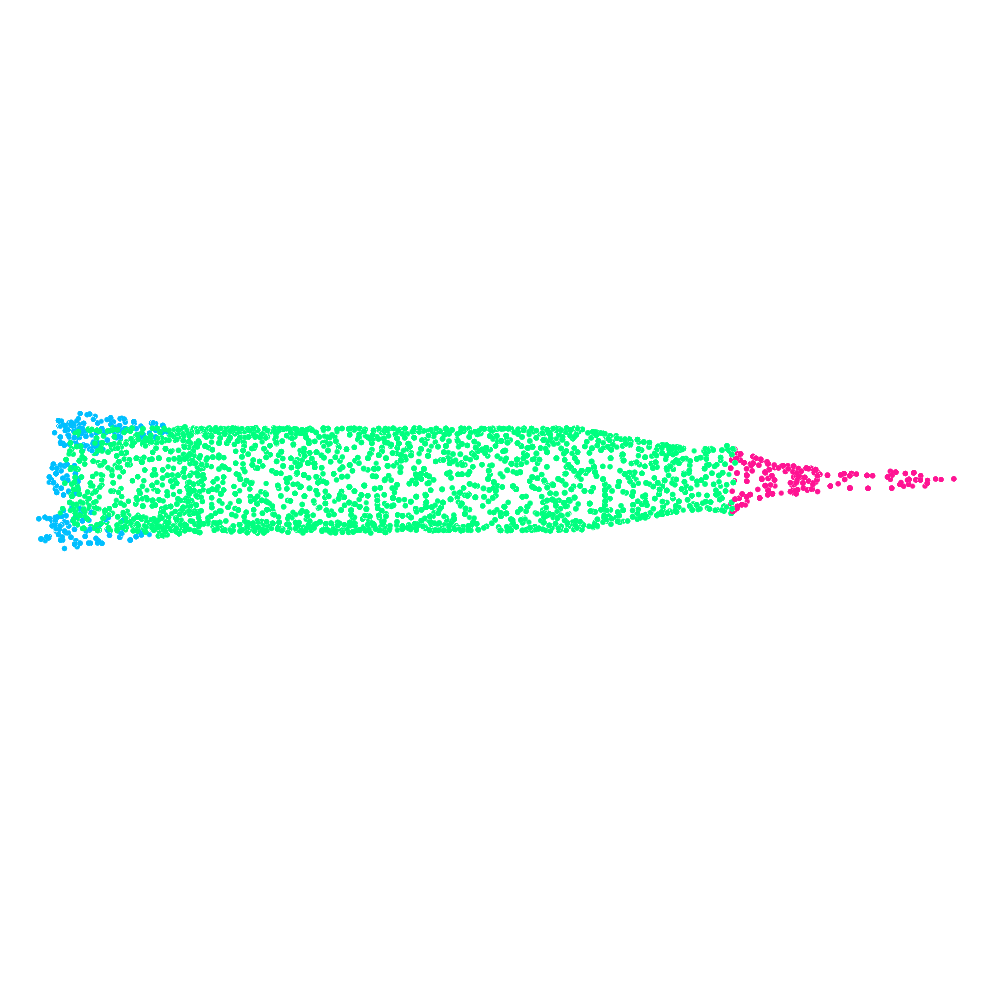}}
        \end{minipage}
        \begin{minipage}[b]{0.47\linewidth}
        {\label{}\includegraphics[width=1\linewidth]{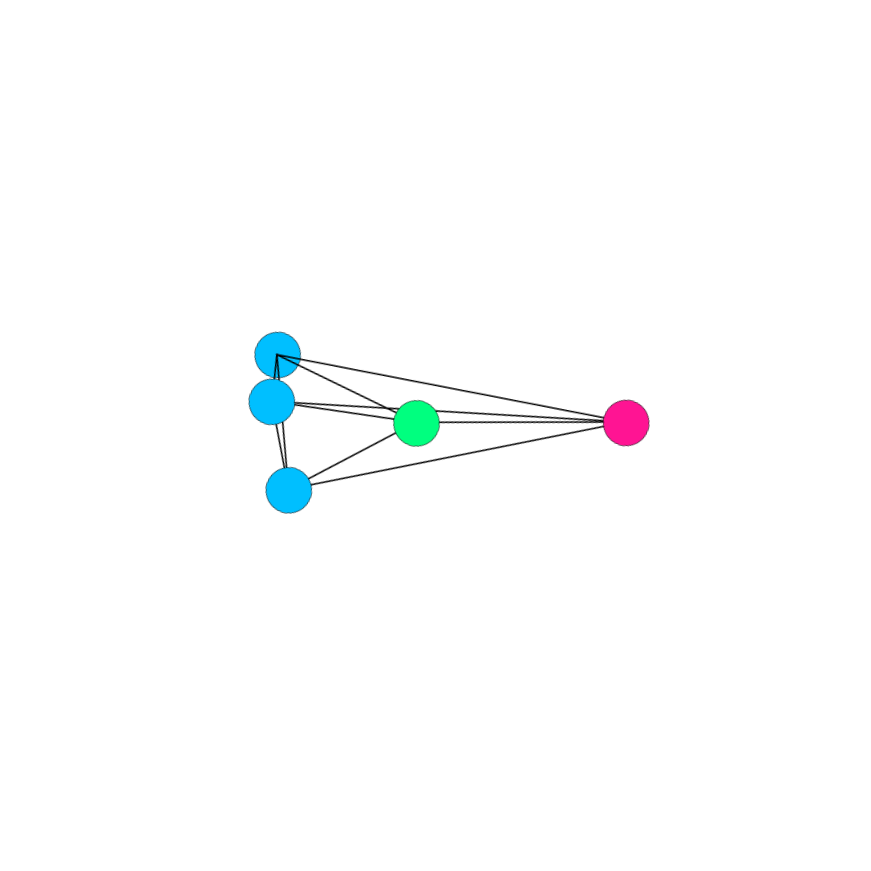}}
        \end{minipage}
    \centerline{\scriptsize{Rocket}}
    \end{minipage}
\end{minipage}
\begin{minipage}[b]{0.49\linewidth}
    \begin{minipage}[b]{0.47\linewidth}
        \begin{minipage}[b]{0.47\linewidth}
        {\label{}\includegraphics[width=1\linewidth]{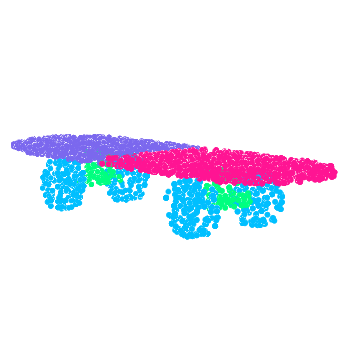}}
        \end{minipage}
        \begin{minipage}[b]{0.47\linewidth}
        {\label{}\includegraphics[width=1\linewidth]{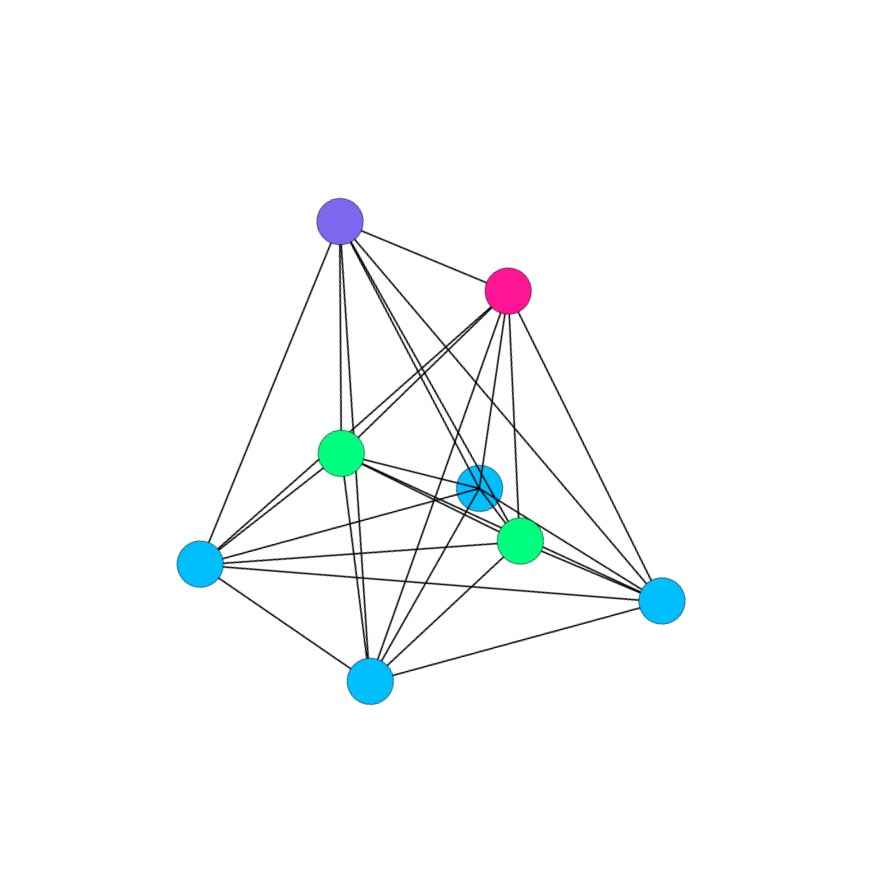}}
        \end{minipage}
    \centerline{\scriptsize{Skateboard}}
    \end{minipage}
    \begin{minipage}[b]{0.47\linewidth}
        \begin{minipage}[b]{0.47\linewidth}
        {\label{}\includegraphics[width=1\linewidth]{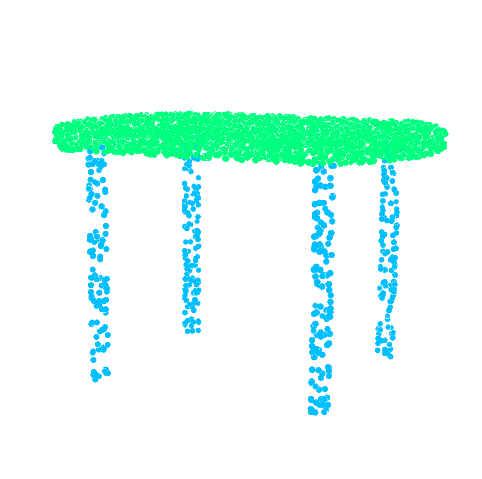}}
        \end{minipage}
        \begin{minipage}[b]{0.47\linewidth}
        {\label{}\includegraphics[width=1\linewidth]{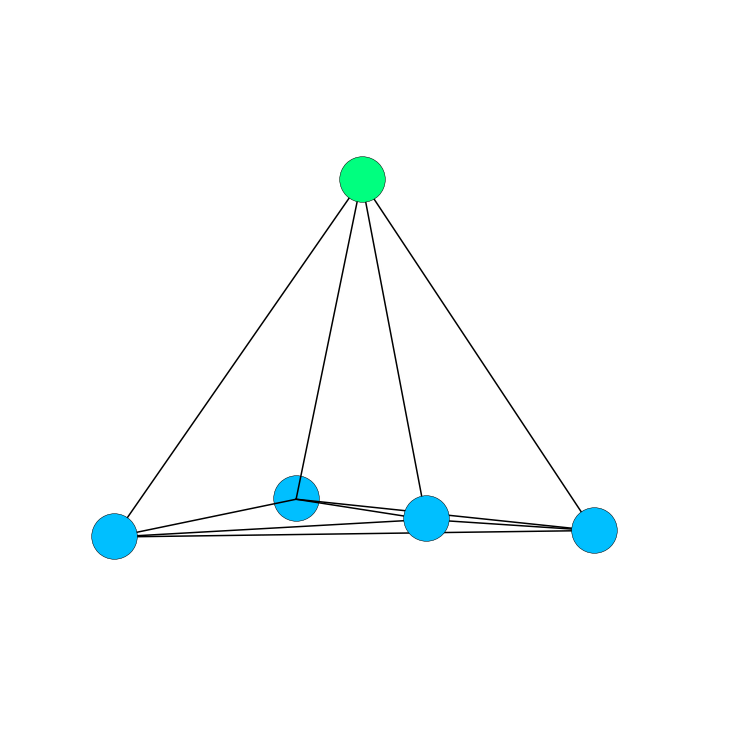}}
        \end{minipage}
    \centerline{\scriptsize{Table}}
    \end{minipage}
\end{minipage}
\\
\begin{minipage}[b]{0.49\linewidth}
    \begin{minipage}[b]{0.47\linewidth}
        \begin{minipage}[b]{0.47\linewidth}
        {\label{}\includegraphics[width=1\linewidth]{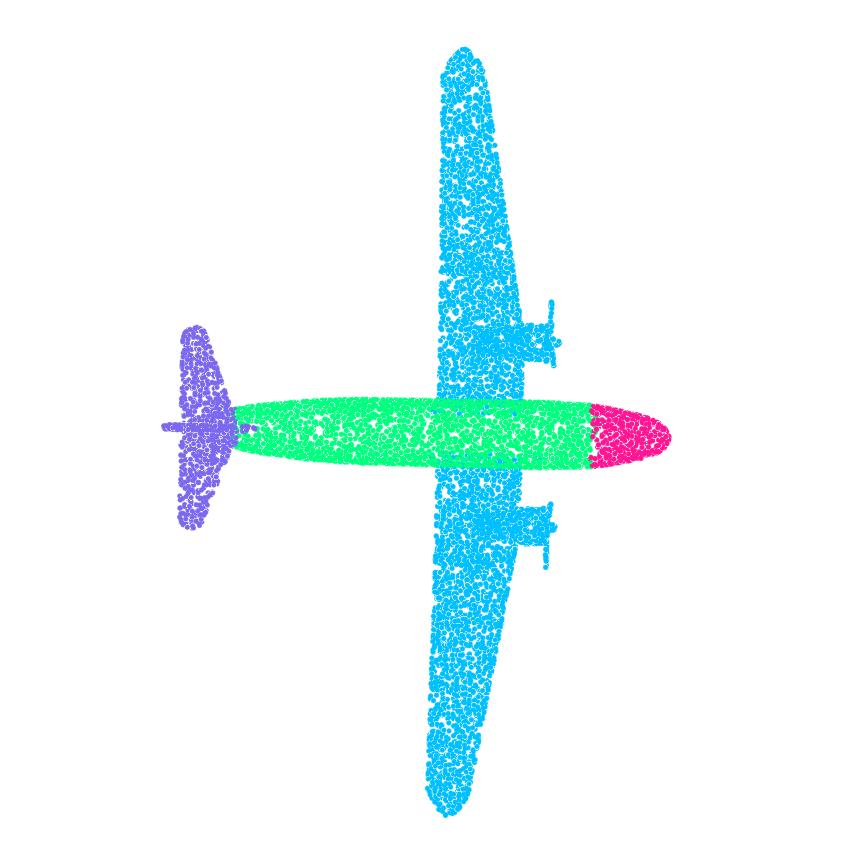}}
        \end{minipage}
        \begin{minipage}[b]{0.47\linewidth}
        {\label{}\includegraphics[width=1\linewidth]{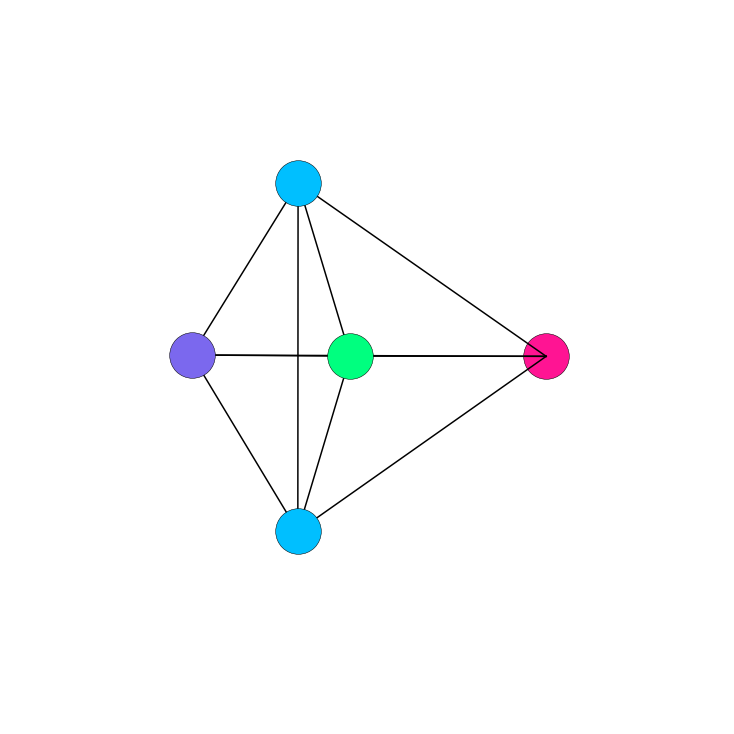}}
        \end{minipage}
    \centerline{\scriptsize{Airplane}}
    \end{minipage}
    \begin{minipage}[b]{0.47\linewidth}
        \begin{minipage}[b]{0.47\linewidth}
        {\label{}\includegraphics[width=1\linewidth]{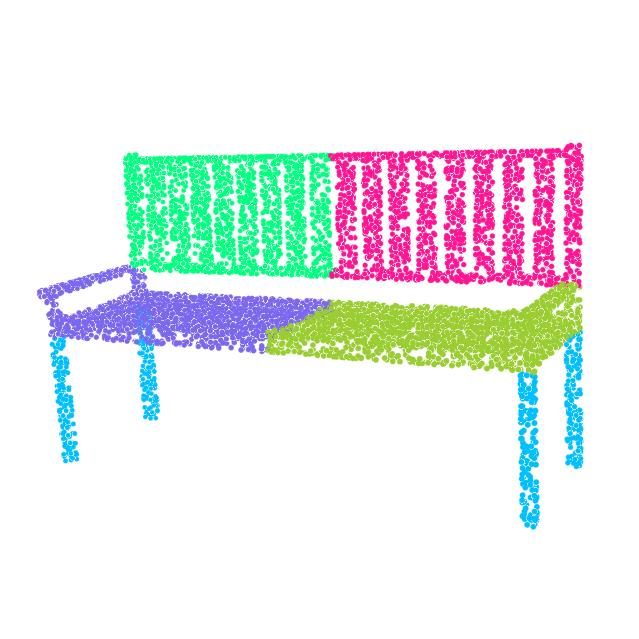}}
        \end{minipage}
        \begin{minipage}[b]{0.47\linewidth}
        {\label{}\includegraphics[width=1\linewidth]{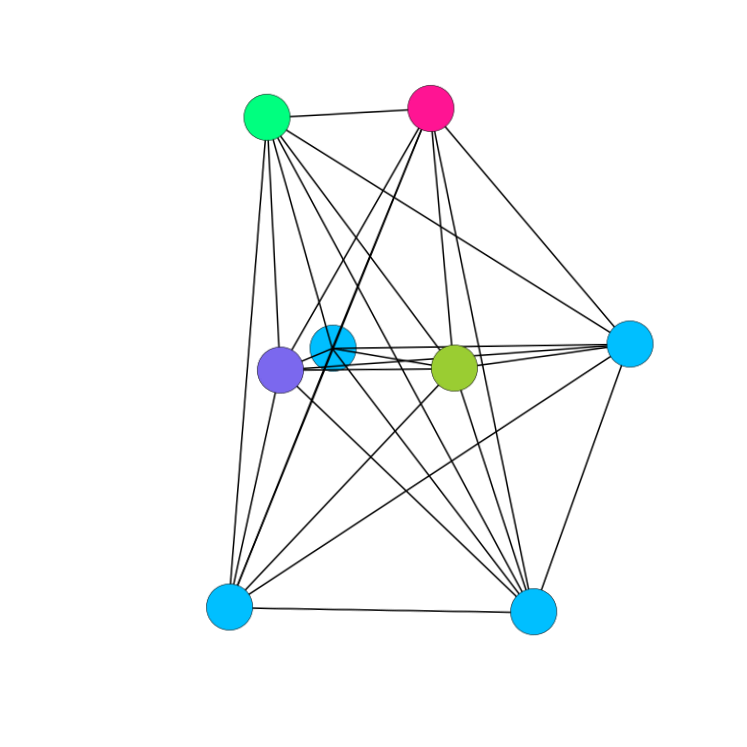}}
        \end{minipage}
    \centerline{\scriptsize{Bench}}
    \end{minipage}
\end{minipage}
\begin{minipage}[b]{0.49\linewidth}
    \begin{minipage}[b]{0.47\linewidth}
        \begin{minipage}[b]{0.47\linewidth}
        {\label{}\includegraphics[width=1\linewidth]{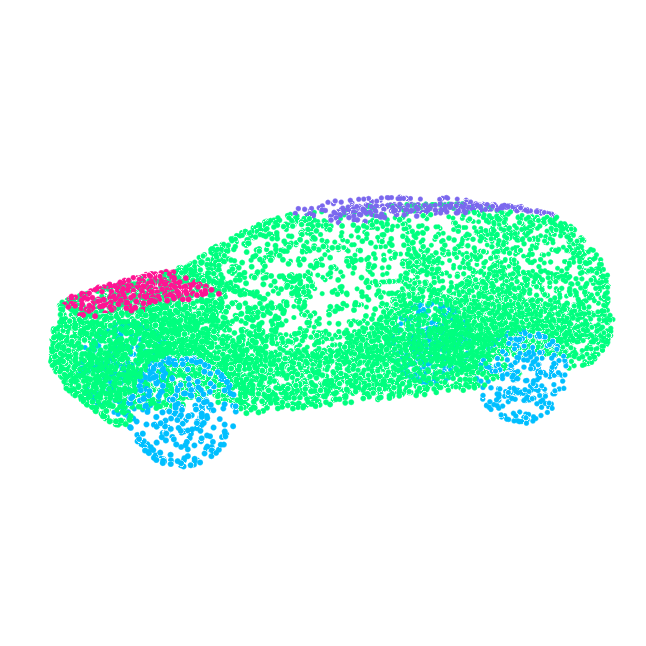}}
        \end{minipage}
        \begin{minipage}[b]{0.47\linewidth}
        {\label{}\includegraphics[width=1\linewidth]{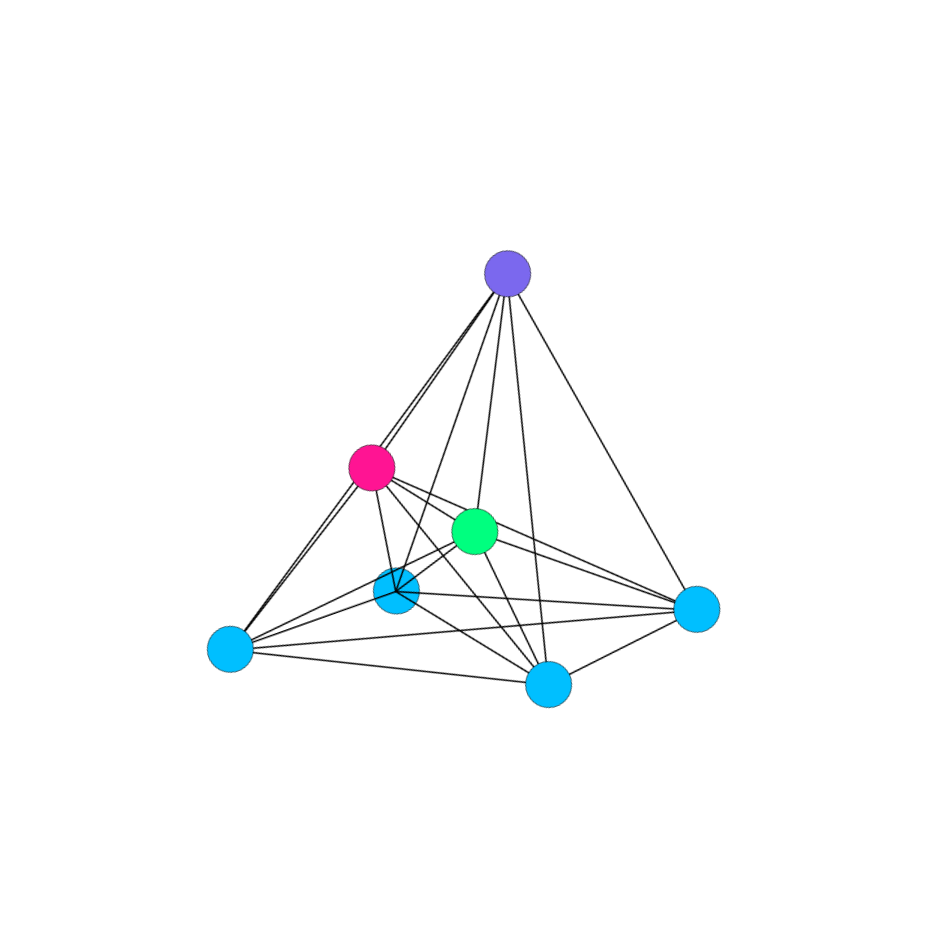}}
        \end{minipage}
    \centerline{\scriptsize{Car}}
    \end{minipage}
    \begin{minipage}[b]{0.47\linewidth}
        \begin{minipage}[b]{0.47\linewidth}
        {\label{}\includegraphics[width=1\linewidth]{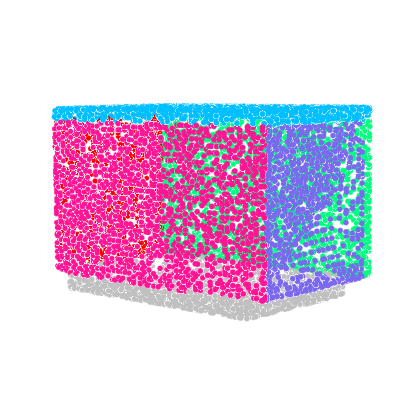}}
        \end{minipage}
        \begin{minipage}[b]{0.47\linewidth}
        {\label{}\includegraphics[width=1\linewidth]{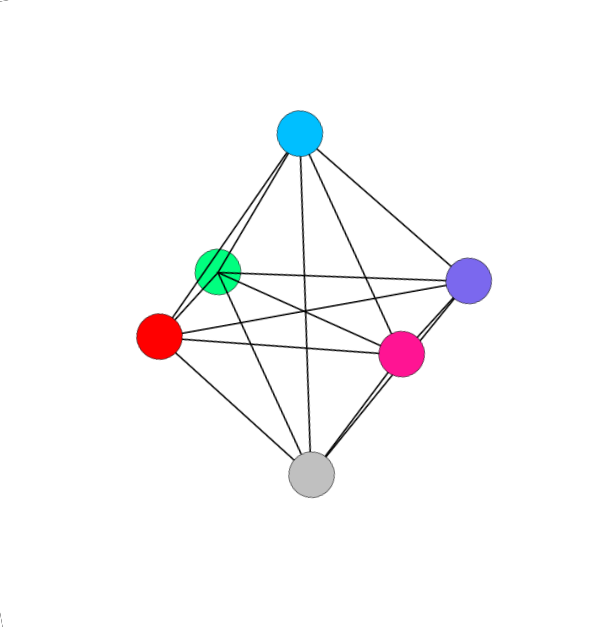}}
        \end{minipage}
    \centerline{\scriptsize{Dresser}}
    \end{minipage}
\end{minipage}
\\
\begin{minipage}[b]{0.49\linewidth}
    \begin{minipage}[b]{0.47\linewidth}
        \begin{minipage}[b]{0.47\linewidth}
        {\label{}\includegraphics[width=1\linewidth]{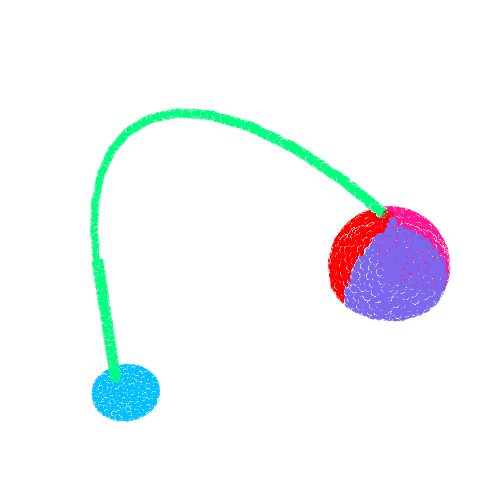}}
        \end{minipage}
        \begin{minipage}[b]{0.47\linewidth}
        {\label{}\includegraphics[width=1\linewidth]{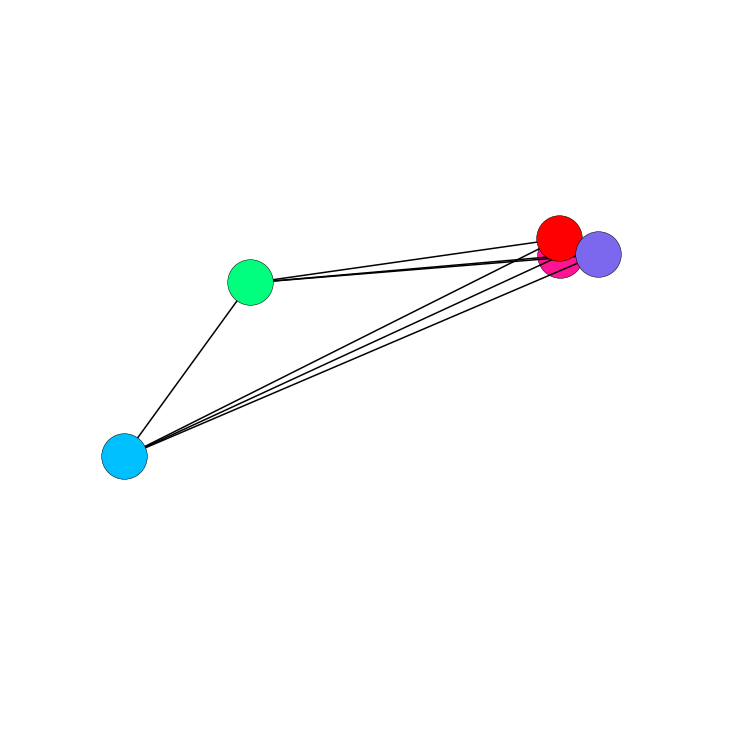}}
        \end{minipage}
    \centerline{\scriptsize{Lamp}}
    \end{minipage}
    \begin{minipage}[b]{0.47\linewidth}
        \begin{minipage}[b]{0.47\linewidth}
        {\label{}\includegraphics[width=1\linewidth]{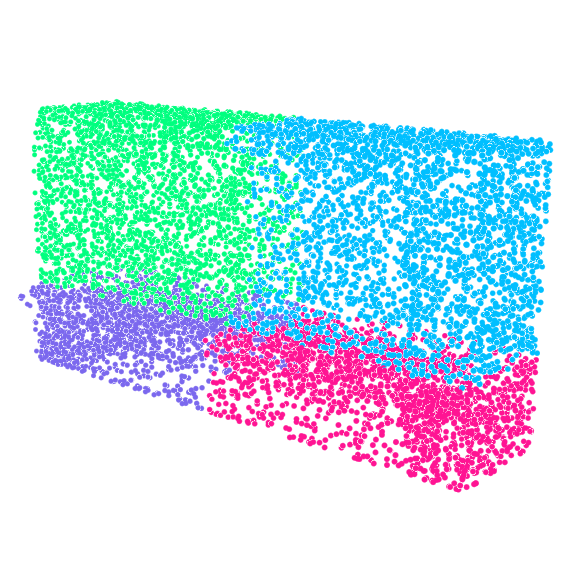}}
        \end{minipage}
        \begin{minipage}[b]{0.47\linewidth}
        {\label{}\includegraphics[width=1\linewidth]{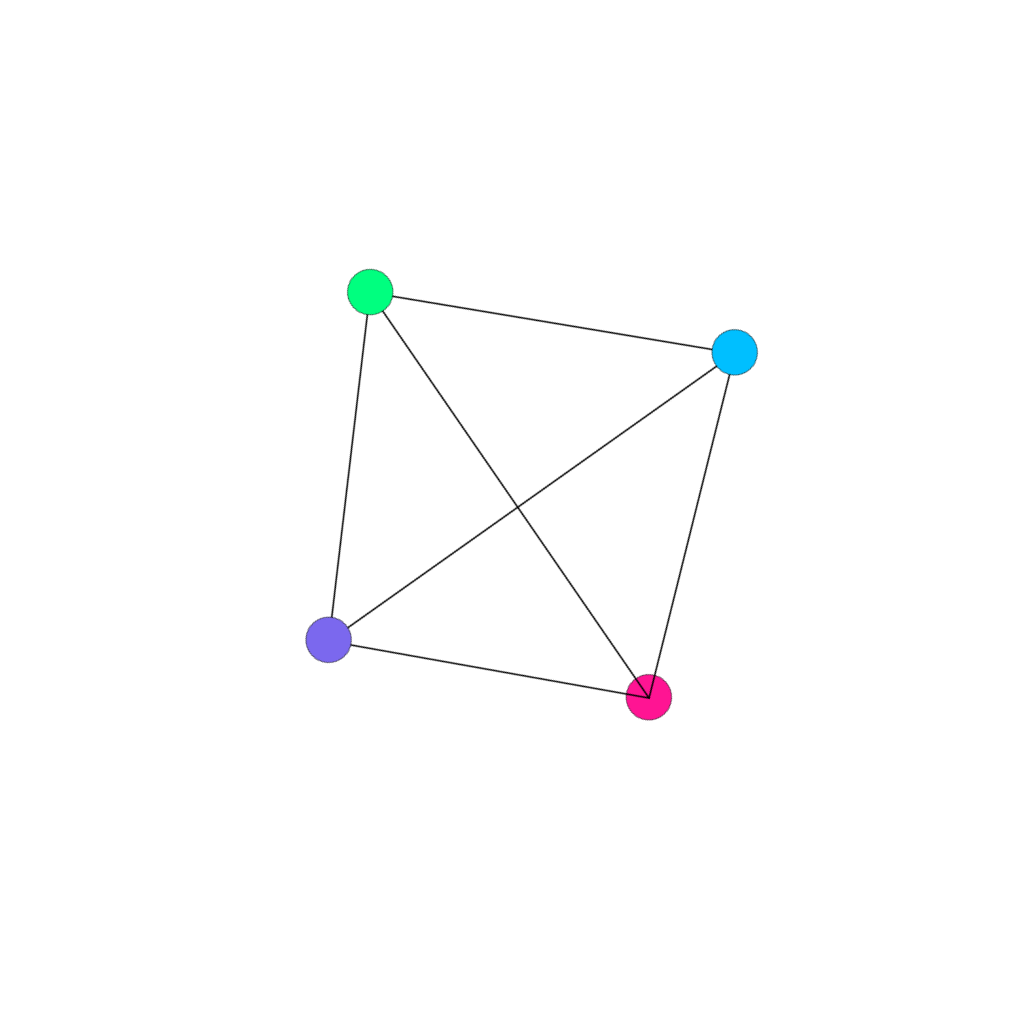}}
        \end{minipage}
    \centerline{\scriptsize{Mantel}}
    \end{minipage}
\end{minipage}
\begin{minipage}[b]{0.49\linewidth}
    \begin{minipage}[b]{0.47\linewidth}
        \begin{minipage}[b]{0.47\linewidth}
        {\label{}\includegraphics[width=1\linewidth]{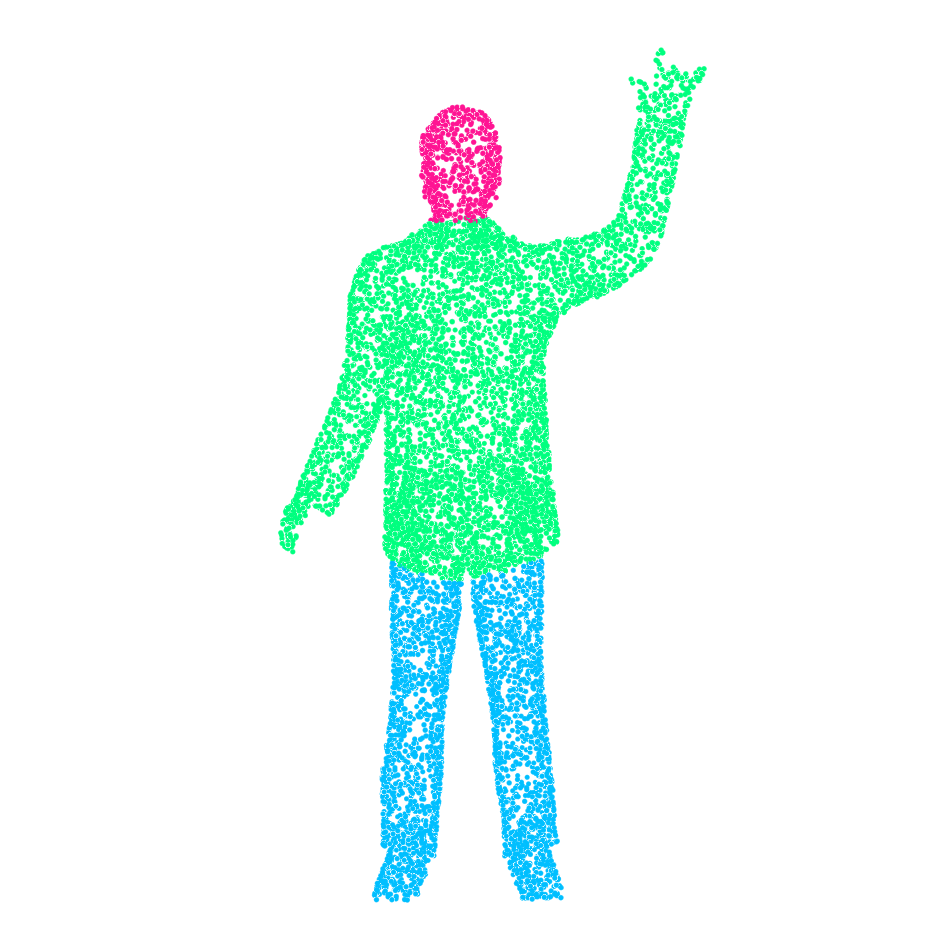}}
        \end{minipage}
        \begin{minipage}[b]{0.47\linewidth}
        {\label{}\includegraphics[width=1\linewidth]{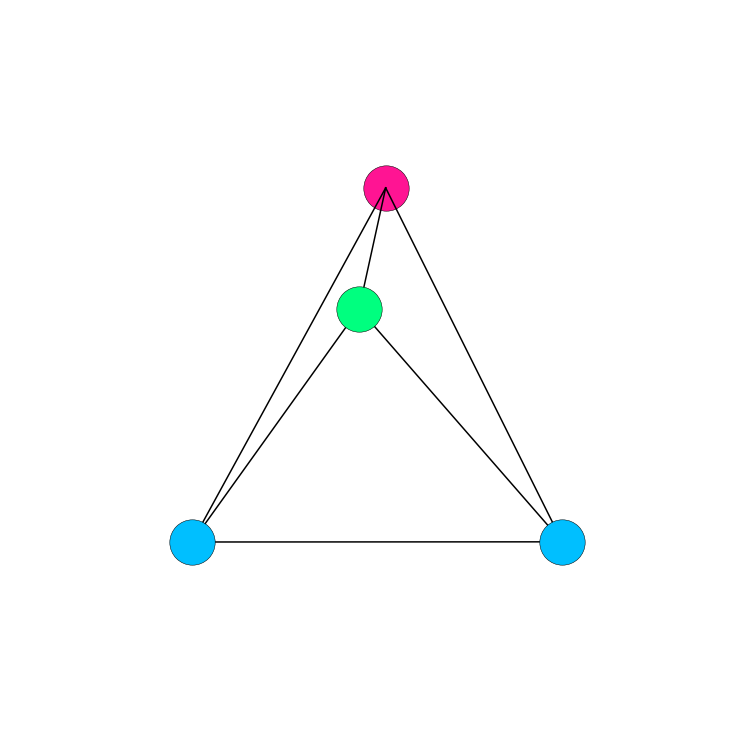}}
        \end{minipage}
    \centerline{\scriptsize{Person}}
    \end{minipage}
    \begin{minipage}[b]{0.47\linewidth}
        \begin{minipage}[b]{0.47\linewidth}
        {\label{}\includegraphics[width=1\linewidth]{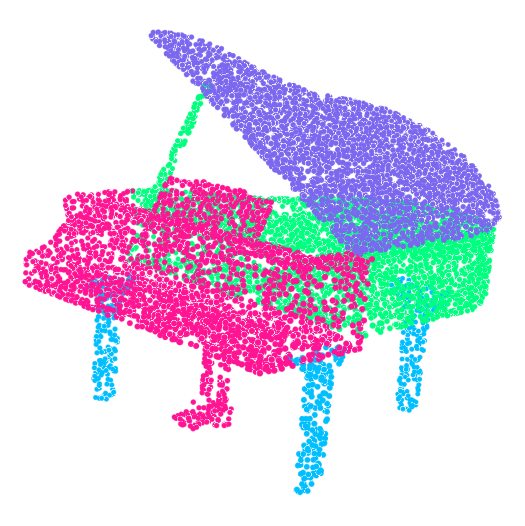}}
        \end{minipage}
        \begin{minipage}[b]{0.47\linewidth}
        {\label{}\includegraphics[width=1\linewidth]{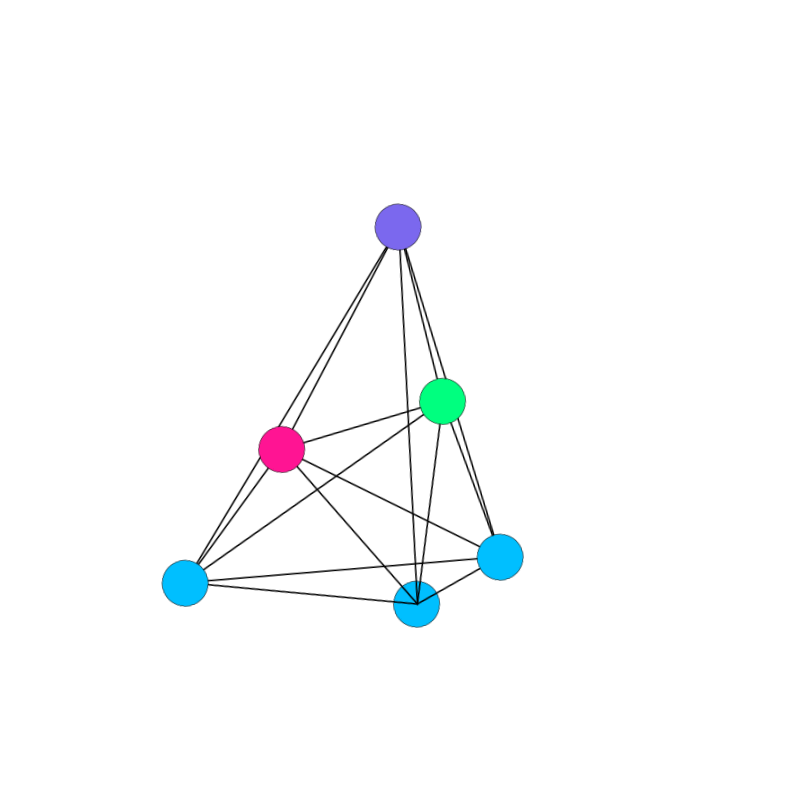}}
        \end{minipage}
    \centerline{\scriptsize{Piano}}
    \end{minipage}
\end{minipage}
\\
\begin{minipage}[b]{0.49\linewidth}
    \begin{minipage}[b]{0.47\linewidth}
        \begin{minipage}[b]{0.47\linewidth}
        {\label{}\includegraphics[width=1\linewidth]{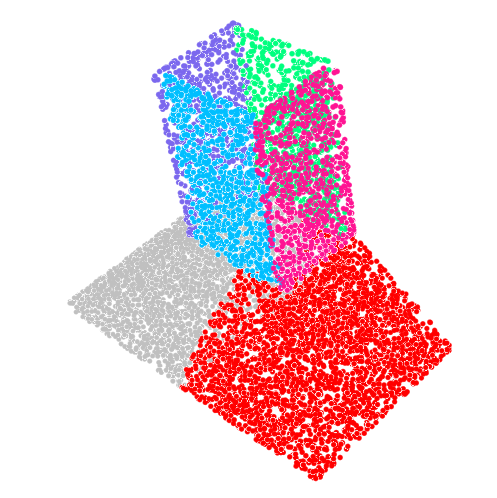}}
        \end{minipage}
        \begin{minipage}[b]{0.47\linewidth}
        {\label{}\includegraphics[width=1\linewidth]{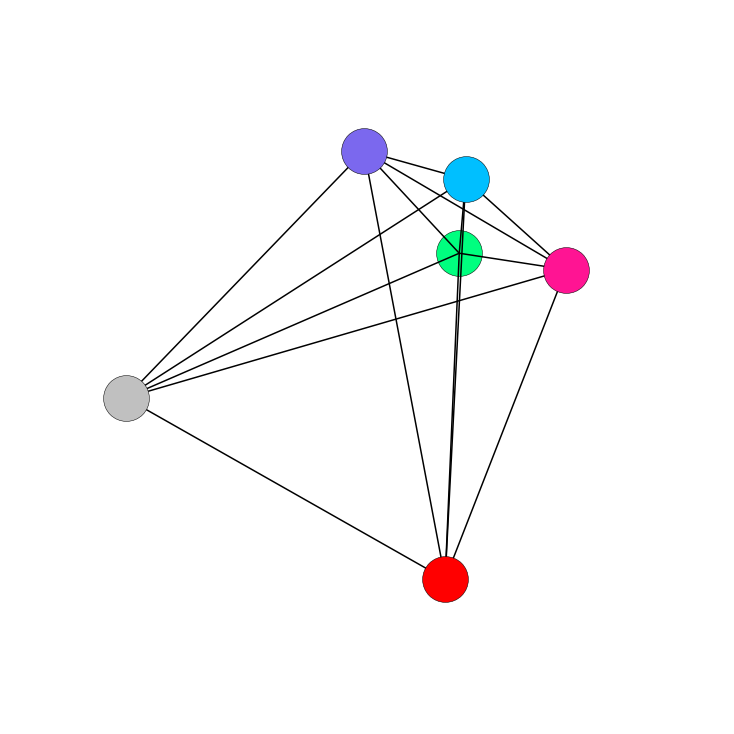}}
        \end{minipage}
    \centerline{\scriptsize{Range Hood}}
    \end{minipage}
    \begin{minipage}[b]{0.47\linewidth}
        \begin{minipage}[b]{0.47\linewidth}
        {\label{}\includegraphics[width=1\linewidth]{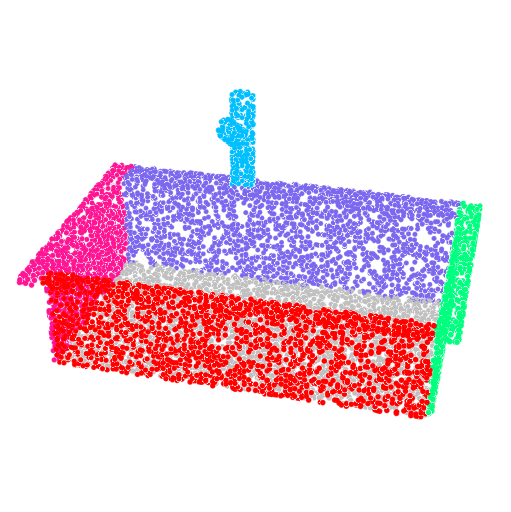}}
        \end{minipage}
        \begin{minipage}[b]{0.47\linewidth}
        {\label{}\includegraphics[width=1\linewidth]{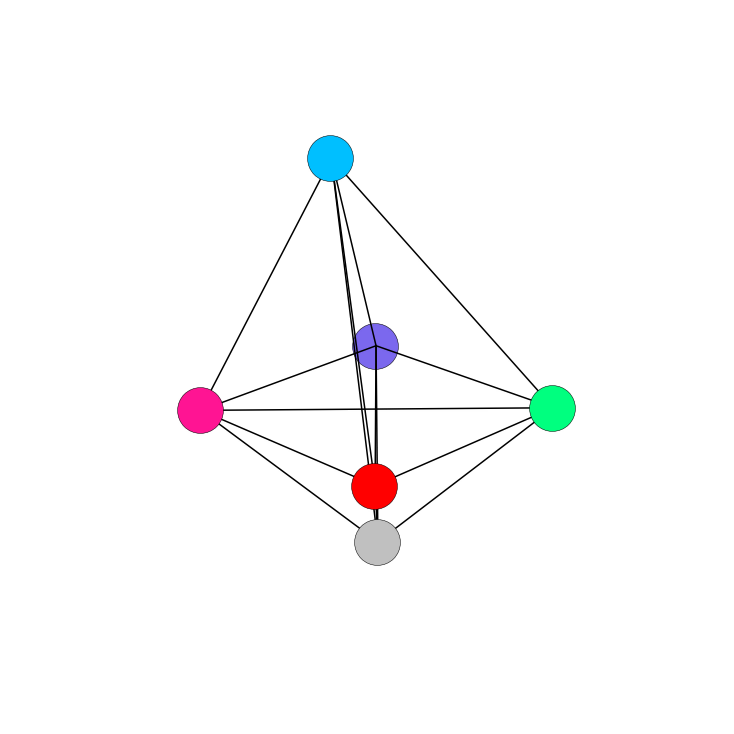}}
        \end{minipage}
    \centerline{\scriptsize{Sink}}
    \end{minipage}
\end{minipage}
\begin{minipage}[b]{0.49\linewidth}
    \begin{minipage}[b]{0.47\linewidth}
        \begin{minipage}[b]{0.47\linewidth}
        {\label{}\includegraphics[width=1\linewidth]{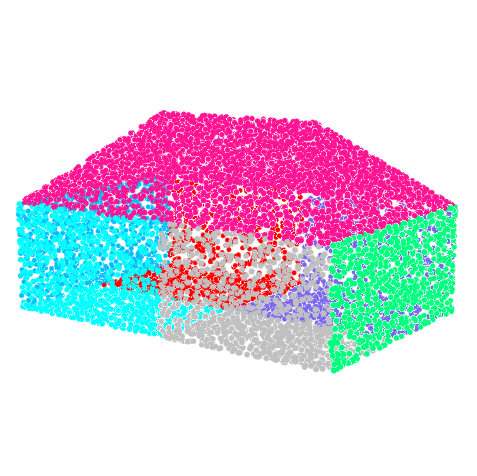}}
        \end{minipage}
        \begin{minipage}[b]{0.47\linewidth}
        {\label{}\includegraphics[width=1\linewidth]{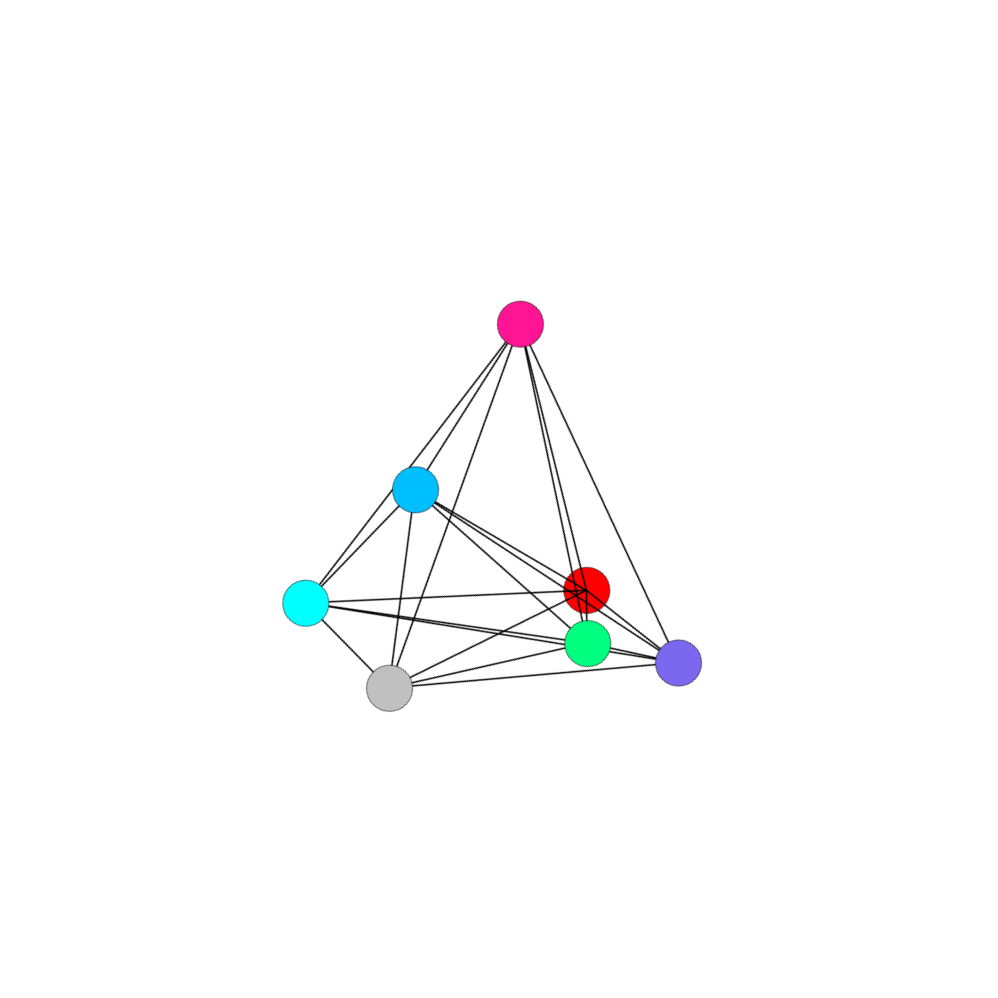}}
        \end{minipage}
    \centerline{\scriptsize{Tent}}
    \end{minipage}
    \begin{minipage}[b]{0.47\linewidth}
        \begin{minipage}[b]{0.47\linewidth}
        {\label{}\includegraphics[width=1\linewidth]{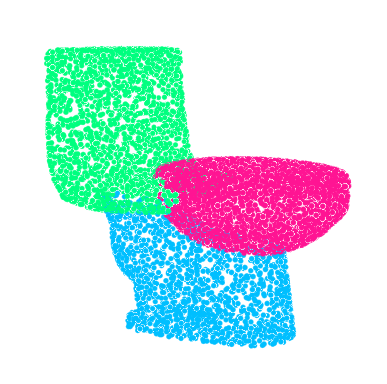}}
        \end{minipage}
        \begin{minipage}[b]{0.47\linewidth}
        {\label{}\includegraphics[width=1\linewidth]{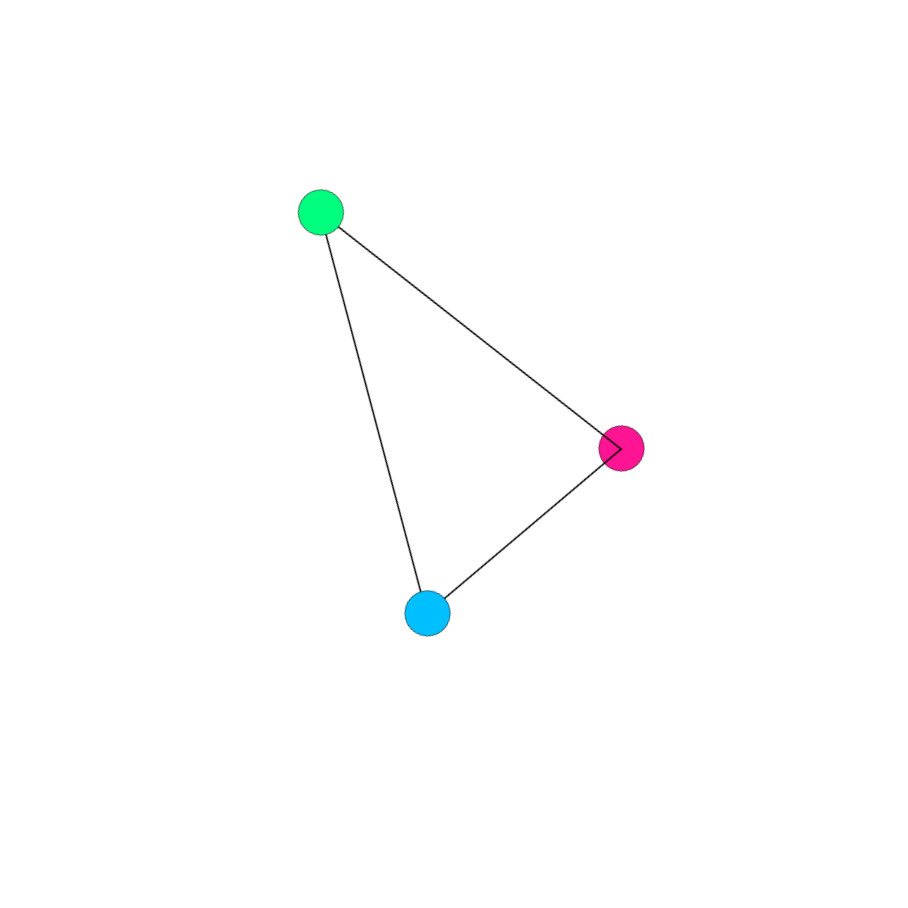}}
        \end{minipage}
    \centerline{\scriptsize{Toilet}}
    \end{minipage}
\end{minipage}
\\
\caption{Samples of segmented point clouds and their corresponding graphs. First two rows:  8 categories from Graph-ShapeNet Part. Last three rows: 12 categories from Graph-ModelNet40. For each category, the same color represents the same semantic part information. Only one edge between two nodes is shown for clarity. }
\label{fig:pc-graph}
\end{figure}

\textbf{Graph-ModelNet40 and Graph-ModelNet10.} ModelNet40 \cite{wu20153d} is the most used dataset for the point cloud classification task. We split the training and testing data following \cite{qi2017pointnet}, i.e., 9,843 point clouds for training and 2,468 for testing. ModelNet10 is a subset of ModelNet40, with 3,991 samples for training and 908 for testing. To reduce the computation time of the ICP algorithm, we select 1,024 points from both the template and its pairing point cloud. In the graph extraction step, the threshold $\tau$ is empirically set to 0.283 as it is usually greater than any distance between two adjacent points, and less than the smallest distance between points in spatially separated parts. Based on ModelNet40 (and its subset ModelNet10) \cite{wu20153d}, we extract two graph datasets, referred to as ``Graph-ModelNet40'' and ``Graph-ModelNet10'', with average 5.3 nodes and 4.6 nodes in each graph. Fig. \ref{fig:pc-graph} shows constructed graphs.

\textbf{Graph-ShapeNet Part dataset.} ShapeNet Part dataset \cite{yi2016scalable} has been frequently used in the point cloud segmentation task due to its abundant categories and accurate part annotations. We transform all the 16 point cloud categories directly to graphs using our method. The created graph dataset is referred to as the ``Graph-ShapeNet Part''. Our constructed  Graph-ShapeNet Part dataset has 4 nodes per graph on average across all the categories. Some constructed graphs are shown in Fig. \ref{fig:pc-graph}.

\subsection{Experimental Settings}
\label{sec4:p2:Settings}

Following \cite{zhang2018end}, we set the user-defined parameter $z$ in the SortPooling layer to the default value 30. For a fair comparison, we train our model and the comparative methods in Tab. \ref{tab:results on constructed graph datasets} for 200 epochs on our graph datasets. The batch size of all the methods is set to 20 for training and testing. Both our model and the backbone model \cite{zhang2018end} are implemented on Tensorflow and trained on an off-the-shelf desktop computer with an Intel Core i7-10750H \textbf{CPU} (rather than GPU). Adam optimizer with a learning rate of 0.001 and the cross-entropy loss are adopted for training.

\subsection{Classification Results on Constructed Datasets}
\label{sec4:p3:classify results 1} 

To validate the proposed method, we firstly compare it with state-of-the-art graph classification methods on our constructed graph datasets. Tab. \ref{tab:results on constructed graph datasets} shows the results, where ``OA'' represents the overall accuracy (mean accuracy of all instances). We adopt the edge features calculated by Eq. \eqref{eq:+x} while implementing the default value as 1 in other methods. It can be seen that our method soundly outperforms state-of-the-art ones on all the three graph datasets, confirming the effectiveness of our graph classification method.

Specifically, our method surpasses the $2^{\text{nd}}$ best method SAGPool \cite{lee2019self} by 0.49\% in terms of OA on Graph-ModelNet40, and exceeds the $2^{\text{nd}}$ best method \cite{diehl2019towards} by 1.42\% in terms of OA on Graph-ShapeNet Part. It is interesting to see our model respectively takes about 1 second and 2 seconds per epoch when training on Graph-ModelNet10 and Graph-ModelNet40 on \textbf{CPU}, which are nearly identical to the training time per epoch for the backbone model \cite{zhang2018end}. This verifies that our add-on layer requires unnoticeable additional overhead. We attribute this energy-efficient computing to the lightweight add-on layers, each of which only introduces 99 parameters to the total 56,779 parameters of the backbone model \cite{zhang2018end}. It is worth noting that methods \cite{baek2021accurate,bianchi2020spectral,papp2021dropgnn} do not perform very well. It might be because these methods are not applicable to our graph datasets that contain only a small number of nodes per graph.

\begin{table}
\footnotesize
\begin{center}
\caption{Comparisons of overall accuracy of ours and state-of-the-art graph classification methods on our graph datasets. G-M40: Graph-ModelNet40, G-M10: Graph-ModelNet10, G-S: Graph-ShapeNet Part. }
\label{tab:results on constructed graph datasets}
\begin{tabular}{lcccc}
\hline\noalign{\smallskip}
Methods & Source & G-M40(OA)  & G-M10(OA) &  G-S(OA)   \\ 
\noalign{\smallskip}
\hline
\noalign{\smallskip}
EdgePool \cite{diehl2019towards} & ICML'19 & 95.42\%& 98.57\% & 96.49\% \\  
SAGPool \cite{lee2019self} & ICML'19 & 96.07\%& 98.57\% & 96.17\% \\ 
MinCUTPool \cite{bianchi2020spectral} & ICML'20 & 81.00\%& 88.22\% & 78.01\%\\
ASAP \cite{ranjan2020asap} & AAAI'20 & 95.95\%& 98.68\% & 96.28\% \\
GMT \cite{baek2021accurate} & ICLR'21 & 80.96\%& 88.77\% & 78.78\% \\
SSRead \cite{lee2021learnable} & ICDM'21 & 95.75\%& 98.46\% & 95.65\% \\
DropGNN \cite{papp2021dropgnn} & NIPS'21 & 85.05\%& 94.93\% & 86.17\% \\
\noalign{\smallskip}    
\hline
\noalign{\smallskip}
Ours & - & \textbf{96.56\%}& \textbf{98.79\%} & \textbf{97.91\%} \\ 
\noalign{\smallskip}
\hline
\noalign{\smallskip}
\end{tabular}
\end{center}
\end{table}

\subsection{Classification Results on Existing Graph Datasets}
\label{sec4:p4:classify results 2}

To test the generalization of our method on existing graph datasets, four commonly used graph datasets (PROTEINS, DD, NCI1 and NCI109) are selected for comparison. The node features of graphs in these datasets are in the format of one-hot encoded labels, which lack discriminations when used in our edge feature scheme for edge feature calculation. Therefore, we take the default 1 as the input edge feature that aligns with other state-of-the-art methods. Following the same settings as \cite{zhang2018end}, we evaluate the performance of all methods with 10-fold cross-validation. Tab. \ref{tab:results on existing graph datasets} shows that our proposed method achieves the best averaged overall accuracy results on all datasets, confirming the generalization of the proposed method in classifying graphs with richer nodes.

\begin{table}
\footnotesize
\begin{center}
\caption{Comparisons of overall accuracy of ours and state-of-the-art graph classification methods on our graph datasets. G-M40: Graph-ModelNet40, G-M10: Graph-ModelNet10, G-S: Graph-ShapeNet Part. }
\label{tab:results on existing graph datasets}
\begin{tabular}{lccccc}
\hline\noalign{\smallskip}
Methods & Source & PROTEINS  &  DD &  NCI1 & NCI109  \\ 
\noalign{\smallskip}
\hline
\noalign{\smallskip}
EdgePool \cite{diehl2019towards} & ICML'19 & $71.50\pm4.70$ & - & - &  - \\  
SAGPool \cite{lee2019self} & ICML'19 & $71.86\pm0.97$ & $76.45\pm0.97$ & $74.18\pm1.20$ & $74.06\pm0.78$  \\ 
MinCUTPool \cite{bianchi2020spectral} & ICML'20 & $76.50\pm2.60$ & $80.80\pm2.30$ & - & -  \\
ASAP \cite{ranjan2020asap} & AAAI'20 & $74.19\pm0.79$ & $76.87\pm0.70$ & $71.48\pm0.42$ & $70.07\pm0.55$  \\
GMT \cite{baek2021accurate} & ICLR'21 & $75.09\pm0.59$ & $78.72\pm0.59$ & - & -  \\
SSRead \cite{lee2021learnable} & ICDM'21 & $69.02\pm1.08$ & $75.86\pm0.61$ & $73.19\pm0.83$ & -  \\
DropGNN \cite{papp2021dropgnn} & NIPS'21 & $77.20\pm4.70$ & - & - & -  \\
\noalign{\smallskip}    
\hline
\noalign{\smallskip}
Ours & - & $\textbf{79.46}\pm6.13$ & $\textbf{81.03}\pm5.82$ & $\textbf{80.03}\pm1.93$ & $\textbf{78.59}\pm4.08$  \\
\noalign{\smallskip}
\hline
\noalign{\smallskip}
\end{tabular}
\end{center}
\end{table}

\subsection{Ablation Study}
\subsubsection{Effectiveness of our edge feature schemes and add-on layers}
\label{sec4:p5:effectiveness}
In this set of experiments, we examine the effectiveness of our 
edge feature schemes and add-on layers in enhancing the graph classification performance on our graph datasets. Here, we take the backbone model \cite{zhang2018end} as baseline and abbreviate it as $BM$, our edge feature schemes (in Sec. \ref{sec3:p2.2:Edge features}) as $EFS$, and our add-on layers (in Sec. \ref{sec3:p2.3:add-on layer}) as $AoL$. 

We build three versions of models for comparison: (1) the backbone model takes the default 1 as the edge feature value, and the model is trained without the proposed add-on layers. (2) The backbone model adopts our edge feature scheme but without the add-on layers. (3) The backbone model adopts both the proposed edge feature scheme and the add-on layers, i.e., ours. 

Tab. \ref{tab:effectiveness} shows the comparison results. By comparing the first version with the second, it can be seen that our computed edge features are more effective than the default value in the graph convolution layer. The results of the second and the third show that our add-on layers enhance the backbone model's feature learning ability, improving the overall accuracy by $1.32\%$ on Graph-ShapeNet Part, $0.66\%$ on Graph-ModelNet10 and $0.81\%$ on Graph-ModelNet40. These results prove the effectiveness of each component devised in our method. 

\begin{table}\footnotesize
\begin{center}
\caption{Comparisons of overall accuracy of different models. BM: backbone model, EFS: edge feature schemes, AoL: add-on layers.}
\label{tab:effectiveness}
\begin{tabular}{c c c c c c}
\hline\noalign{\smallskip} 
BM & EFS & AoL & G-S& G-M10 & G-M40\\ 
\noalign{\smallskip}
\hline
\noalign{\smallskip}
 \checkmark & & & 95.37\% & 98.02\% & 95.26\%  \\
 \checkmark &\checkmark& & 96.59\% & 98.13\% & 95.75\%  \\
  \checkmark&\checkmark&\checkmark& \textbf{97.91}\% & \textbf{98.79}\% & \textbf{96.56}\% \\
\noalign{\smallskip}
\hline
\noalign{\smallskip}
\end{tabular}
\end{center}
\end{table}

\subsubsection{Edge features}
\label{sec4:p5:edge features}

Unlike the edge feature scheme defined in Eq. \eqref{eq:+x}, a traditional Gaussian kernel used to define edge features in adjacency matrix is like below.

\begin{equation}\label{eq:gaussian kernel}
\begin{aligned}
\tilde{\mathbf{A}}^{k}_{w,q} = e^{-\frac{\left\|\mathrm{H}^{k}_{{v}_{w}} - \mathrm{H}^{k}_{{v}_{q}}\right\|_{2}^{2}}{\sigma^2}},
\end{aligned}
\end{equation}
where $\sigma$ is the kernel width that is usually set to a constant. This formula controls the edge features between 0 to 1, and gets a higher edge feature value when the square of the $L_{2}\mhyphen norm$ of each two node representations is lower. 

On our graph datasets, we conduct experiments to compare the performance of the above edge feature schemes without altering add-on layers. Also, we include the default 1 for edge features and a positive exponent in Eq. \eqref{eq:gaussian kernel}. For a fair comparison, $\sigma$ is set to 1. It can be seen from Tab. \ref{tab:edge features} that using traditional Gaussian kernel in the edge feature schemes may lead to less accurate topological information being encoded, thus resulting in lower accuracy. By contrast, our edge feature schemes encode better topological information and lead to higher accuracy.

\begin{table}
\begin{center}
\caption{Comparisons of overall accuracy of different edge feature update schemes.}
\label{tab:edge features}
\begin{tabular}{c c c c}
\hline\noalign{\smallskip}
Edge feature schemes & 
G-S&
G-M10&
G-M40\\ 
\noalign{\smallskip}
\hline
\noalign{\smallskip}
$Default$  & 97.46\% &  98.24\% & 95.95\% \\

$e^{-\left\|\mathrm{H}^{k}_{{v}_{w}} - \mathrm{H}^{k}_{{v}_{q}}\right\|_{2}^{2}}$ & 96.63\% & 96.26\% & 93.92\%  \\

$e^{\left\|\mathrm{H}^{k}_{{v}_{w}} - \mathrm{H}^{k}_{{v}_{q}}\right\|_{2}^{2}}$ & 97.81\% & 98.13\% & 95.87\%  \\

$e^{\left\|\mathrm{H}^{k}_{{v}_{w}} - \mathrm{H}^{k}_{{v}_{q}}\right\|_{2}}$ (Ours) & \textbf{97.91}\% & \textbf{98.79}\% & \textbf{96.56}\% \\
\noalign{\smallskip}
\hline
\noalign{\smallskip}
\end{tabular}
\end{center}
\end{table}

\subsubsection{Normalized adjacency matrix}
\label{sec4:p5:adj matrix}
Normalized adjacency matrix reflects how features are aggregated on graphs. Some methods \cite{te2018rgcnn,li2018adaptive,zhang2019hierarchical,li2019semi} chose the symmetric normalized matrix as the normalization method in \cite{bruna2013spectral}. We conduct experiments to compare the graph classification accuracy with four types of normalized adjacency matrices, i.e., column normalized matrix $\tilde{\mathbf{A}}\tilde{\mathbf{D}}^{-1}$, naively symmetric normalized matrix $\tilde{\mathbf{D}}^{-1} \tilde{\mathbf{A}} \tilde{\mathbf{D}}^{-1}$, symmetrically normalized matrix $\tilde{\mathbf{D}}^{-\frac{1}{2}} \tilde{\mathbf{A}} \tilde{\mathbf{D}}^{-\frac{1}{2}}$ and our row normalized matrix $\tilde{\mathbf{D}}^{-1} \tilde{\mathbf{A}}$, where $\tilde{\mathbf{A}}$ is the same as that in Eq. \eqref{eq:+x}. 

We implement these four normalized adjacency matrices in our model and train each of them on our graph datasets. The comparisons of the overall accuracy are listed in Tab. \ref{tab: normalization}. It can be seen that the row normalized matrix has a better performance over the other three on all the graph datasets. 

\begin{table}
\begin{center}
\caption{Comparisons of overall accuracy of different adjacency normalization methods in adjacency matrix. 
}
\label{tab: normalization}
\begin{tabular}{c c c c}
\hline
\noalign{\smallskip}
Normalization Method & 
G-S&
G-M10&
G-M40\\ 
\noalign{\smallskip}
\hline
\noalign{\smallskip}
$\tilde{\mathbf{A}}\tilde{\mathbf{D}}^{-1}$ &  97.84\% &  98.46\% & 96.31\%      \\
$\tilde{\mathbf{D}}^{-1} \tilde{\mathbf{A}} \tilde{\mathbf{D}}^{-1}$ & 97.70\% & 97.47\% & 96.43\%  \\
$\tilde{\mathbf{D}}^{-\frac{1}{2}} \tilde{\mathbf{A}} \tilde{\mathbf{D}}^{-\frac{1}{2}}$ & 97.63\% & 98.35\% & 96.31\%  \\
$\tilde{\mathbf{D}}^{-1} \tilde{\mathbf{A}}$ (Ours) & \textbf{97.91}\% & \textbf{98.79}\% & \textbf{96.56}\% \\
\noalign{\smallskip}
\hline
\end{tabular}
\end{center}
\end{table}

\subsection{Limitations and Future Work}
\label{ch4:p6:future work}

Since our edge feature schemes generate edge features through node features, some datasets include their own edge features as input which are not compatible with our method, e.g., MUTAG and PTC are two datasets in which edge features are labels the edges belong to. In future, we would like to design a more general edge feature scheme and add-on layer that is considering of those initial edge labels.

%% file: paper/conclusion.tex
\section{Conclusion}
\label{sec5:conclusion}
In this paper, we introduce a graph classification method by deeply exploiting the node and edge features of graphs. An edge feature scheme and an add-on layer are designed to optimize graph structures for effective graph learning in spectral GCNNs. A graph construction method is introduced which generates graph datasets for method validation. Extensive experiments on the constructed Graph-ModelNet40, Graph-ModelNet10 and Graph-ShapeNet Part datasets demonstrate that our method outperforms state-of-the-art graph classification techniques.